%% file: main.tex
\newif\ifamcs
\amcsfalse %

\newif\ifincludeappendix
\includeappendixtrue

\ifamcs
    \documentclass{amcs}
\else
    \documentclass{article}
    \usepackage{arxiv}
\fi

\usepackage[utf8]{inputenc} %
\usepackage[T1]{fontenc}    %
\usepackage{hyperref}       %
\usepackage{url}            %
\usepackage{booktabs}       %
\usepackage{amsfonts}       %
\usepackage{nicefrac}       %
\usepackage{microtype}      %
\usepackage{lipsum}
\usepackage{amsmath}
\usepackage{float}
\usepackage{xfrac}
\usepackage{multirow} 

\usepackage{array}
\newcolumntype{H}{>{\setbox0=\hbox\bgroup}c<{\egroup}@{}}

\ifamcs
    \usepackage{times}
    \usepackage{amssymb}
    \usepackage{color}
    \usepackage[hang]{caption2}
\fi

\title{Learning Abstract Visual Reasoning via Task Decomposition: A Case Study in Raven Progressive Matrices}

\ifamcs
    \author[ad1][]{Jakub KWIATKOWSKI} %
    \author[ad1][]{Krzysztof KRAWIEC}
    \correspondingauthor{Jakub KWIATKOWSKI}
    \address[ad1]{Institute of Computing Science\\
    Poznan University of Te/nology\\
    Piotrowo 2, 60-965 Poznań, Poland\\
    e-mail: \url{jakub.k.kwiatkowski@doctorate.put.poznan.pl, krawiec@cs.put.poznan.pl}
    }
    \Runauthors{J. Kwiatkowski and K. Krawiec}

\else
\author{
    Jakub Kwiatkowski
    \\
    Institute of Computing Science
    \\
    Poznan University of Technology
    \\
    \texttt{jakub.k.kwiatkowski@doctorate.put.poznan.pl}
    \\
    \And
    Krzysztof Krawiec
    \\
    Institute of Computing Science
    \\
    Poznan University of Technology
    \\
    \texttt{krawiec@cs.put.poznan.pl}
    \\
}
\fi

\begin{document}

\ifamcs\else
\maketitle
\fi

\begin{abstract}
    Learning to perform abstract reasoning often requires decomposing the task in question into intermediate subgoals that are not specified upfront, but need to be autonomously devised by the learner. In Raven Progressive Matrices (RPM), the task is to choose one of the available answers given a context, where both the context and answers are composite images featuring multiple objects in various spatial arrangements. As this high-level goal is the only guidance available, learning to solve RPMs is challenging. %
    In this study, we propose a deep learning architecture based on the transformer blueprint which, rather than directly making the above choice, addresses the subgoal of predicting the visual properties of individual objects and their arrangements. The multidimensional predictions obtained in this way are then directly juxtaposed to choose the answer. We consider a few ways in which the model parses the visual input into tokens and several regimes of masking parts of the input in self-supervised training. In experimental assessment, the models not only outperform state-of-the-art methods but also provide interesting insights and partial explanations about the inference. The design of the method also makes it immune to biases that are known to be present in some RPM benchmarks.  
\end{abstract}

    \hypertarget{introduction}{\section{Introduction}\label{introduction}}

\begin{figure}
        \ifamcs
\else
        \centerline{\includegraphics[width=0.5\columnwidth]{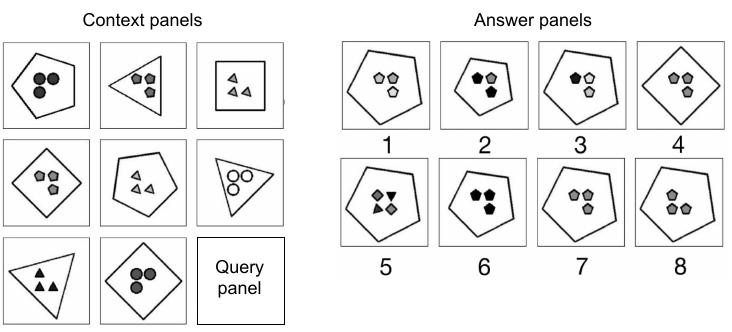}}
        \fi
        \caption{An example of an RPM task.}\label{fig:RPM}
    \end{figure}

    One of the key attributes of general intelligence is abstract reasoning, which, among others, subsumes the capacity to reason about and complete sequential patterns. To quantify such capabilities in humans and diagnose related deficiencies, John C. Raven devised in the 1930s a visual test contemporarily known as Raven Progressive Matrices (RPM) \cite{raven1936mental}. An RPM problem (Fig.~\ref{fig:RPM}) is a 3x3 grid of eight \emph{context panels} containing arrangements of 2D geometric objects. The objects adhere to rules that govern the relationships between the panels in rows, e.g. progression of the number of objects, a logical operation concerning their presence, etc. The task is to complete the puzzle by replacing the lower-right \emph{query panel} with the correct image from the eight provided \emph{answer panels}. Solving the task requires 'disentangling' the rules corresponding to rows and columns and capturing the analogies between the observed patterns. 
    
    RPMs have been more recently adopted in the AI community for assessing similar capabilities in artificial intelligent systems, along with other benchmarks like Bongard problems \cite{bongard1967problem} and Hofstadter's analogies \cite{Hofstader1995}. Recent advances in machine learning accelerated this trend, with deep learning becoming the primary paradigm of choice for designing such systems \cite{Malkinski_Mandziuk_2022}.

    The original collection of RPM problems, Standard Progressive Matrices  \cite{raven1936mental}, comprised just 60 tasks, which is not enough to effectively train data-hungry machine learning models. Therefore, several larger datasets and task generators have been devised, among them RAVEN \cite{zhang2019raven} and I-RAVEN \cite{Hu2021}. %
    In this process, it became evident that designing a representative, diverse, and varying in difficulty collection of RPM tasks is nontrivial. The key challenge is that one needs to generate 7 incorrect answer panels such that it is impossible to deduce the correct answer panel from them. Unfortunately, most tasks in RAVEN do not meet this requirement: the correct answer panel can be selected by identifying the most frequent attributes across all 8 answer panels (shape, size, etc.) and picking the answer panel with those properties. This flaw can be easily exploited, which was epitomized with so-called \emph{context-blind} methods \cite{Wu2020} that achieve almost perfect scores on RAVEN by disregarding the context entirely and making decisions based on answer panels only. This problem has been termed \emph{biased answer set}, and we illustrate it in Sec.\ 
    \ifincludeappendix
    ~\ref{sec:biased-answer-set}
    \fi  of the Supplementary Material (SM).  

    In this study, we circumvent this problem by decomposing RPM tasks into two stages: 1) prediction of properties of the query panel, and 2) identifying the answer panel with properties that match those predicted ones the most. To this aim, we use the abstract properties available in RAVEN benchmarks and design a bespoke deep learning architecture based on the transformer blueprint \cite{Vaswani2017}. The resulting approach, which we dub Abstract Compositional Transformer, is not only more transparent than end-to-end neural architectures but also immune to biased answer sets and capable of surpassing the state-of-the-art performance. More specifically, the property prediction stage is immune to biases because it does not involve the answer panels, while the second stage does not involve learning and thus by definition cannot be biased by the content of a training set. Also, to the best knowledge of the authors, this is the first attempt to predict symbolic descriptors of RPM puzzles and the first study on self-supervised learning for RPM. The two-stage approach and model architecture (Sec.~\ref{sec:prop-pred}), a bespoke training procedure (Sec.~\ref{sec:training}) and an extensive empirical analysis concerning property prediction (Sec.~\ref{sec:exp-prop}) and problem solving (Sec.~\ref{sec:exp-choice}) form our main contributions.

    \section{The proposed approach}\label{sec:prop-pred}

\begin{figure*}
        \ifamcs
\else
        \centerline{\includegraphics[width=\textwidth]{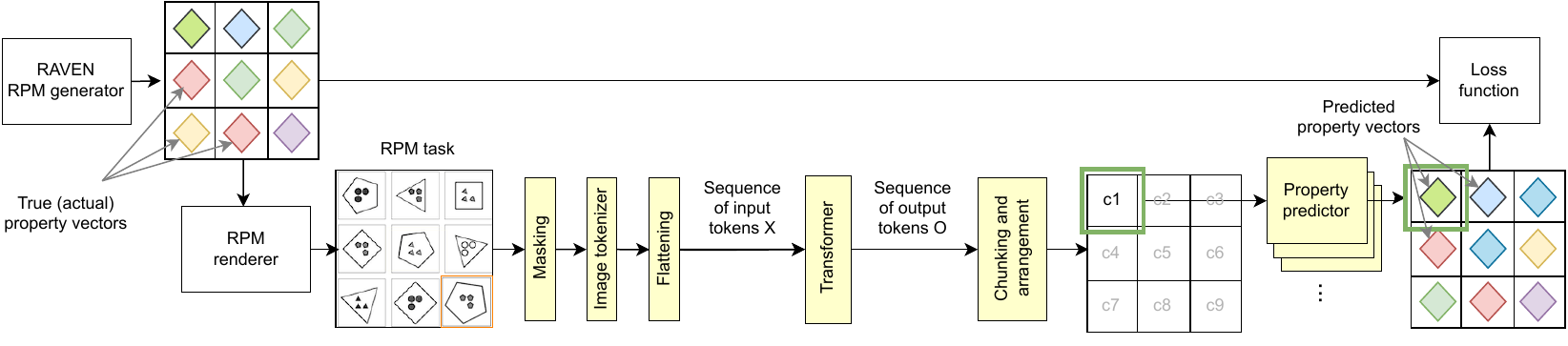}} %
        \fi
        \caption{The architecture of the model (yellow boxes) and its training process, guided by the loss function that compares the predicted and actual properties of RPM panels. The model learns from completed RPM tasks, with one of the panels (context or query panel) masked out, and predicts the properties of all panels. }
        \label{fig:method}
    \end{figure*}

    Rather than training models that choose answer panels in RPM, we propose to rely on \emph{property prediction}, in which models generate an abstract, structured representation of the missing panel (Fig.~\ref{fig:method}). 
    To this aim, we rely on the RAVEN dataset  \cite{zhang2019raven}, in which tasks have been generated from symbolic specifications expressed in an image description grammar that captures visual concepts such as position, type of shape, color, number of objects, inside-outside, etc. The objective of the model is to predict these properties for the query panel and for the answer panels. A trained property prediction model is then subsequently used to choose the answer panel. 

    Our model comprises three modules: image tokenizer, transformer, and property predictor; we describe them in subsections that follow (see
    \ifincludeappendix
    Sec.~\ref{sec:technical-params}
    \fi 
    in the Supplementary Material for technical details). %
    Even though RPM problems have an inherent 2D structure, we rely on \emph{sequence-to-sequence} transformers \cite{Vaswani2017} and demonstrate, analogously to prior work on applying such models to 2D images \cite{Dosovitskiy2020}, that effective RPM solvers can be obtained without explicitly presenting the spatial structure of the problem to the model. %

    \subsection{Image tokenizer}\label{sec:tokenizer}
    
    The tokenizer maps the 2D raster representation of an RPM problem to a sequence of abstract tokens using a convolutional neural network (CNN) that gradually contracts the input image to lower spatial resolutions in consecutive layers, while increasing the number of channels. The multi-channel superpixels produced by the last layer form the tokens, i.e. a token is the vector of channels produced by the CNN at a given location. In experiments, our CNN is the EfficientNetV2B0 %
    \cite{tan2021} pretrained on the ImageNet database \cite{deng2009imagenet}.  
    We consider the three following types of tokenizers that vary in how they perceive the panel rasters (when necessary, the single-channel monochrome RPM image is broadcast to RGB input channels of the CNN). 

    \textbf{Panel tokenizer.}~In this variant, the raster image representing each panel is tokenized independently. The CNN is applied to each raster (84x84 pixels in RAVEN) and produces a 3x3 array of 128-channel superpixels, which is then flattened row-wise into a sequence of nine 128-dimensional tokens. This is repeated for all nine panels of the puzzle, and the resulting sequences are concatenated, producing 81 tokens in total.  

    \textbf{Task tokenizer.}~In this variant, the raster image of the entire RPM task is tokenized with a single invocation of the CNN. For RAVEN, this means applying the CNN to a 252x252 pixel image, which results in an 8x8 array of 128-channel superpixels, then flattened row-wise, leading to a sequence of 64 128-dimensional tokens.  
    
    \textbf{Row tokenizer.}~In this variant, the rasters representing individual panels in each row are first stacked channel-wise, resulting in three 3-channel 84x84 images corresponding to the top, middle, and bottom rows of the puzzle. The CNN is queried on each of those images independently and produces a 3x3 array of superpixels in each query, which are then flattened row-wise, resulting in nine 128-dimensional tokens. Finally, the subsequences for the top, middle, and bottom RPM rows are concatenated, resulting in 27 tokens. 
    
    By stacking the panel images, we directly juxtapose them in input channels, allowing so the CNN to form low-level features that capture the differences between the individual images. The RPM images from the left, central, and right columns end up being interpreted by the CNN as, respectively, the red, green, and blue channels. %
    The pretrained CNN is trained alongside the entire model, and can thus adapt to the characteristics of RPM. \hfill$\square$  

    The relatively small sizes of rasters, combined with 18 convolutional layers of the CNN (cf. Table 1 in \cite{tan2019}), cause the receptive fields of units in the last layer to span the entire input image. Therefore, for all tokenizers, each token may in principle depend on the entire input raster (a panel raster for Panel and Row tokenizers, and a task raster for the Task tokenizer). Also, only the Panel tokenizer is guaranteed to ensure some degree of selective correspondence between RPM panels and tokens. In the Task tokenizer, the representation is more entangled, as the receptive fields of the CNN are allowed to span \emph{multiple} neighboring panels. In the Row tokenizer, the consecutive groups of nine tokens form an entangled representation of the top, middle and bottom row of panels. However, the degree of entanglement depends on the characteristics of features acquired in training, and the actual \emph{effective receptive fields} can be smaller (see, e.g., \cite{Luo2017}).

    \subsection{Sequence-to-sequence transformer}
    
    The transformer processes the one-dimensional sequences of tokens $X$ produced by an image tokenizer by first encoding each token independently using the encoder, 
    \begin{equation}
                    Z = map(Encoder_{\theta_E}, X),
    \end{equation}
    which is implemented as a dense linear layer, primarily meant to match the dimensionality of the tokens to the input dimensionality of the transformer. Then, the transformer maps the sequence of encoded tokens $Z$ to a sequence of output tokens $O$ of the same length:
    \begin{equation}
            O = Transformer(Z; \theta_T).
    \end{equation}
    The model is equipped with a \emph{learnable positional encoding}, applied to the input tokens in $Z$. %
    As the number of tokens is constant, we encode the \emph{absolute} positions of tokens in $Z$, which can be achieved with a fixed-size learnable embedding. There is a single entry in the embedding for each position in the input sequence, and thus 81, 64, and 27  entries for respectively the Panel, Task, and Row tokenizer. The embedding vectors are added to respective tokens in $Z$ before passing them to the transformer.  

    Internally, the transformer is a stack of transformer blocks, each of them consisting of a multi-head attention mechanism $Attn(\theta_A)$, normalization layers $LayerNorm$, a feed-forward network $f(\theta_f)$ and skip connections. The processing for the $i$th token can be summarized as:
    \begin{equation}
        \begin{split}
            a_i & = Attn(LayerNorm(z_i; \theta_N); \theta_A)\\
            m_i & = a_i + z_i \\
            o_i & = f(LayerNorm(m_i; \theta_{N'}); \theta_f) + m_i 
        \end{split}
    \end{equation}
    The resulting tokens $o_i$ form the output sequence $O$. For a detailed description of transformer and multi-head attention, see \cite{Vaswani2017}. %
    
    \subsection{Property predictor}\label{sec:prop-predictor} 

    The property predictor maps the sequence of tokens $O$ produced by the transformer to the properties of individual panels, as follows:
    \vspace{-6pt}\begin{enumerate}
        \item The sequence is sliced into 9 \emph{chunks} of equal length,
        \item The tokens in each chunk are concatenated to form a single vector,
        \item Each vector is independently mapped to a \emph{property vector} (Sec.~\ref{sec:property-vectors}) using a dense subnetwork. 
    \end{enumerate}\vspace{-6pt}
    For the Panel tokenizer, which produces 81 tokens, the chunk length is 9; for the Row tokenizer, which produces 27 tokens, the chunk length is 3; for the Task tokenizer, producing 64 tokens, each chunk comprises 7 tokens, and the last token is discarded.
    
    The nine property vectors obtained in this way are assumed to correspond to RPM panels, traversed row-wise and left-to-right in each row (8 context panels and the query panel). Associating the chunks with panels requires the transformer to both \emph{combine and disentangle} the information carried by the input tokens. The combining is necessitated by the task, which requires detecting the patterns adhered to by the context panels. The disentanglement, on the other hand, is necessary for the Task and Row tokenizers, which do not derive tokens from individual context panels \emph{independently}, but aggregate information from multiple panels. %

    \subsection{Property vectors}\label{sec:property-vectors} 
    
    Following the RAVEN family of benchmarks \cite{zhang2019raven,Hu2021,Benny2021}, we assume the panels to be composed of \emph{objects} that can appear in one of 7 \emph{spatial arrangements}, each containing at least one object, with the maximum number of objects as follows:
    \emph{center-single} (1),
    \emph{distribute-four} (4),
    \emph{distribute-nine} (9),
    \emph{in-center-single-out-center-single} (2),
    \emph{in-distribute-four-out-center-single} (5),
    \emph{left-center-single-right-center-single} (2),
    \emph{up-center-single-down-center-single} (2).
    An RPM panel is represented as a property vector of a fixed dimensionality comprising `slots' for all objects in every  arrangement; there are thus 25 slots in total. %
    Each object is characterized by three \emph{appearance properties} with the following admissible values:  
    \vspace{-6pt}\begin{itemize}
    \item color: 255, 224, 196, 168, 140, 112, 84, 56, 28, 0 (10 values, rendered as colors in this paper),
    \item size: 0.4, 0.5, 0.6, 0.7, 0.8, 0.9 (6 values),
    \item type: triangle, square, pentagon, hexagon, circle (5 values). %
    \end{itemize}\vspace{-6pt}
    Another appearance property used in \cite{zhang2019raven} is object's rotation angle. However, the angle is drawn at random when generating a panel, i.e. it does not influence the reasoning rules.  A successful solver should disregard this characteristic, hence we do not include it in our representation.
    
    In total, a property vector comprises thus 101 variables:
    \vspace{-6pt}\begin{itemize}
        \item the identifier of the arrangement (1 variable), 
        \item \emph{present}$_i$: a group of 25 binary variables, each indicating the presence/absence in the $i$th object slot,
        \item 75 appearance properties for the objects in all slots.
    \end{itemize}\vspace{-6pt}

    \textbf{Relevance of properties}. 
    An element of a property vector may deem some other elements \emph{relevant} or \emph{irrelevant}. For instance, if \emph{arrangement} = \emph{distribute-four}, only \emph{present}$_i$ for four indices $i$ matter; if then only two of them are set to 1, only the corresponding $2\times3=6$ appearance properties describing the two indicated objects are relevant. The number of relevant properties varies from 5 for \emph{center-single} (1 \emph{arrangement} + \emph{present}$_i$ for a single object + 3 appearance properties) to 37 for \emph{distribute-nine} (1 \emph{arrangement} + 9 \emph{present}$_i$ properties + $9\times3$ appearance properties).  

    The distinction between relevant and irrelevant panel properties is essential; in particular, the loss function and the metrics are calculated from relevant properties only. When confronting two property vectors, one of them serves as the \emph{source of relevance}; this depends on the use case (more details in Sec. \ifincludeappendix
    ~\ref{sec:relevance-sources}
    \fi
    and in the experimental part).

    \textbf{Encoding of property vectors}. 
    To make properties amenable to differentiable loss functions, we represent all variables with one-hot-encoding, which results in the low-level representation of a panel being a binary vector of $7+25+25\times(10+6+5)= 557$ dimensions. Respectively, models produce for each panel a $557$-dimensional vector with values in $[0,1]$, which is assured by forcing the dimensions representing a given categorical variable through the softmax activation function; for instance, the first 7 dimensions represent the probability distribution over arrangements.          
    To calculate the loss, the distribution predicted by the model is confronted with the corresponding one-hot target distribution using categorical entropy. The entropies obtained so for particular variables are multiplied by weights tuned and fixed in the early stages of this project (see 
    \ifincludeappendix
    Sec.~\ref{sec:loss-function}
    \fi 
    of the Supplementary Material). %
    The loss is calculated only for the distributions of relevant properties, with the target property vector acting as the source of relevance. %
    The overall loss on a given RPM task is a weighted sum of the losses for individual panels, where weights depend on the type of masking applied in training (Sec.~\ref{sec:training}).

    \section{Model training}\label{sec:training}

    As in natural language processing, our models are trained via self-supervision, i.e. they are tasked to predict a masked-out element of the input sequence, given the context of the visible elements. While masking usually concerns tokens, RPMs require making decisions about panels, %
    therefore in each training step, the tokenizer is applied to the task with a single context panel masked out. %
    
    In RPM, one can make predictions in both directions of the sequence of context panels, due to the nature of the underlying patterns (e.g. the progression of the number of objects). It seems thus desirable to mask randomly chosen panels to facilitate learning of patterns \emph{across the entire RPM puzzle}, and prospectively use that knowledge for making decisions about the query panel. The query panel is empty by definition, so it seems natural to treat it as a masked-out one too. However, masking out more than one panel at once would be inconsistent with test-time querying (when only the query panel is masked) and could lead to ambiguity, i.e.~multiple answer panels being correct. Therefore, we split training into two phases: %
    \begin{enumerate}
        \item \textbf{Random masking phase}. Each puzzle is completed with the correct answer from the answer panels, and a randomly chosen panel (one of 9) is masked out. This is realized `on the fly', so the same task has different panels masked out in different epochs. %
        \item \textbf{Query masking phase}. The RPM tasks are presented as-is, with the query panel masked out, and no additional masking is applied. The weight of the loss related to the query panel is multiplied by 0.01, in order for this phase to act as fine-tuning after Phase 1. The loss function is not applied to the non-masked panels, i.e.~the model is not penalized for making predictions there.   
    \end{enumerate}
    By staging training into these phases, we allow the model to first learn the patterns across the entire RPM board, and only then require it to focus on the query panel. 
    
    Masking requires replacing a panel with a 'neutral' image; initially, we considered empty (zeroed) panels and random noise images. Ultimately, the \emph{trainable masks} performed best: the masking image is initialized with random is treated as a parameter of the model, i.e.~it is updated in training, ultimately expressing the cumulative input that the model `expects' at masked panels.

    \hypertarget{related-work}{\section{Related work}\label{related-work}}

    RPM has been long considered an interesting benchmark for abstract reasoning systems, along with Bongard problems \cite{bongard1967problem}, Hofstadter's analogies \cite{Hofstader1995}, Numbo \cite{Defays1995}, or Sudoku, to name a few. 
    The advent of deep learning only intensified this interest, with an outpour of studies proposing various architectures and learning approaches. A recent survey \cite{Malkinski_Mandziuk_2022} %
    cites at least 34 papers, most of them published within the last five years; Tables \ref{tab:sota-raven} and \ref{tab:sota-iraven} cite those of them that achieved best performances on RAVEN \cite{zhang2019raven} and I-RAVEN \cite{Hu2021} benchmarks. Of those, the model that bears the most architectural similarity to the approach proposed in this paper is the Attention Relation Network (ARNe) \cite{Hahne2019}, which engages a transformer to facilitate spatial abstract reasoning. However, like almost all other studies that the authors of this study are aware of, ARNe is trained to choose answer panels, rather than to predict panel properties, and thus uses the transformer blueprint in a very different way.  

    In terms of the taxonomy proposed in \cite{Malkinski_Mandziuk_2022}, our approach could be classified as a relational reasoning network, as a part of the model is delegated to learn relations between context panels. A notable representative of this class is the Wild Relation Network (WReN) \cite{barrett2018}, %
    in which a relation network is used to score the answer panels. 
    In contrast, our approach does not model the relations between panels explicitly, but delegates relational learning to the transformer, while encoding the spatial characteristics of the task as a sequence of tokens. 
    
    Our approach bears resemblance to some past works in the hierarchical networks category delineated in \cite{Malkinski_Mandziuk_2022}. More specifically, the way in which our Row tokenizer stacks panels channel-wise is analogous to the design of `perceptual backbones' in, e.g., the Stratified Rule-Aware Network (SRAN; \cite{Hu2021};  %
    see also Fig.~8 in \cite{Malkinski_Mandziuk_2022}). Notice, however, that the panel stacking used in the Row tokenizer is the only way in which we explicitly reveal the relationships between panels to the model. All remaining logic about the correspondence, succession, progression, etc.~of patterns in the panels needs to be autonomously learned by juxtaposing input tokens. This mitigates manual modeling of relationships and, consequently, human biases. 

    Last but not least, there were several works in which the models, apart from choosing the right answer panel, were required to make predictions about the \emph{rules} that govern the generation of RPM tasks. This has been attempted via auxiliary terms in the loss function in \cite{zhang2019raven}, but the models were not benefitting from this extension, or even underperforming when compared to the reference architecture (Sec.~6.4 therein). Similar negative results have been reported in, among others, \cite{Hu2021}, \cite{Zhang2019} %
    and several follow-up studies (see Sec.~3.2 in  \cite{Malkinski_Mandziuk_2022}). Preliminary encouraging results in addressing these challenges have been presented in \cite{Malkinski2022}.  %

    \section{Results: property prediction}\label{sec:exp-prop}

    In this section, we cover the experimental results for the property prediction tasks; in Sec.~\ref{sec:exp-choice}, we use the trained models to solve the choice tasks. Implementation details are available in the Supplementary Material (SM). %

    Following \cite{zhang2019raven}, we use the original division of the $70{,}000$  tasks from the RAVEN database\footnote{\url{https://github.com/WellyZhang/RAVEN}} (7 spatial arrangements~$\times10{,}000$ tasks) into training, validation and test sets of, respectively, $42{,}000$, $14{,}000$, and $14{,}000$ tasks. 
    We train nine models in total, using the three types of image tokenizers and three masking regimes in training (Sec.~\ref{sec:training}): \textbf{Combined} comprising 200 epochs of random masking and up to 30 epochs of query masking, and two ablative regimes: \textbf{Query}-only (query masking for 200 epochs) and \textbf{Random}-only (random masking for 200 epochs). In the first phase of masking (whether followed by a second phase or not), the weight of the loss for the masked panel is multiplied by 2, to emphasize its importance (this multiplier value was found beneficial in preliminary experiments). %
    Validation takes place after each epoch, and the model with the lowest validation error is selected. 
    
    To assess the models' capacity to predict panel properties, we devise a range of test-set metrics that are calculated on the relevant properties only, with the target property vector acting as the source of relevance, i.e. determining the properties that are deemed relevant for a given task (Secs.~\ref{sec:property-vectors} and 
    \ifincludeappendix
    \ref{sec:relevance-sources}):
    \fi   
    \begin{itemize}
        \item \textbf{Correct}: The primary metric in further considerations. Amounts to 1 if \emph{all} relevant properties of the query panel have been correctly predicted; otherwise 0. Averaged across all tasks in the testing set.  %
        \item \textbf{PropRate}: The fraction of correctly predicted relevant properties, across all test tasks. 
        \item \textbf{AvgProp}: The fraction of relevant properties correctly predicted, averaged over tasks. For a given task, it amounts to 1 if all relevant properties have been predicted correctly, and 0 if none. Because the number of relevant properties varies by task and panel, AvgProp is not equivalent to PropRate. %
        \item \textbf{AvgH}: The Hamming distance between the predictions and the target on the relevant properties, averaged over tasks. For a given task, the best attainable value of this metric is obviously 0, while the worst one corresponds to the scenario with 9 objects (\emph{distribute-nine} arrangement) and amounts to 37 (1 for the incorrect identifier of the arrangement, plus 9 for the incorrect setting of the 9 corresponding presence/absence variables, plus $9\times3=27$ incorrect values of the color, size and type of an object).  %
    \end{itemize}

    \textbf{Results}. In the \emph{Property prediction} part of Table \ref{tab:tokenizers-masking}, we report the performance of particular models. %
    Both the type of tokenizer and the masking scheme are strong determinants of model capabilities. The models that tokenize each panel separately (Panel) fare the worst on all metrics. 
    Tokenizing the entire task at once (Task) leads to much better predictive accuracy. Nevertheless, the model that involves the Row tokenizer systematically fares best when juxtaposed with the others, which suggests that superimposing panels as separate image channels facilitates inferring relevant patterns. 

\begin{table*}
\centering
\caption{Comparison of configurations on the property prediction task. } %
\label{tab:tokenizers-masking}
\begin{tabular}{llrrrrrr} %
\toprule
Tokenizer & Masking & \multicolumn{4}{c}{Property prediction} & \multicolumn{2}{c}{Classification} \\
\cmidrule(lr){3-6} \cmidrule(lr){7-8}
         &         & Correct &  PropRate &  AvgProp &  AvgH &  Correct &  AvgProp \\
\midrule
Panel & Query &          1.53 &      62.23 &     58.08 &         4.55 &               97.54 &    99.50 \\
         & Random &         20.82 &      82.52 &     80.34 &         2.23 &           98.38 &    99.44 \\
         & Combined &         22.17 &      83.33 &     81.29 &         2.14 &         98.28 &    99.49 \\ \midrule
Task & Query &         18.91 &      81.23 &     79.25 &         2.41 &                89.69 &    98.35 \\ 
         & Random &         72.63 &      95.80 &     95.25 &         0.65 &           88.44 &    98.00 \\
         & Combined &         75.63 &      96.15 &     95.64 &         0.61 &         88.33 &    97.98 \\ \midrule
Row & Query &         20.56 &      79.96 &     78.15 &         2.66 &                 84.03 &    97.68 \\
         & Random &         75.44 &      96.17 &     95.69 &         0.60 &           87.64 &    98.22 \\
         & Combined &         77.58 &      96.47 &     95.99 &         0.56 &         87.85 &    98.25 \\
\bottomrule
\end{tabular}
\end{table*}

    The observed differences might be partly due to the number of tokens used in particular architectures (81, 64, and 27 for Panel, Task and Row).  %
    However, the Row tokenizer uses the \emph{fewest} tokens, so it is in principle most likely to suffer from the `information bottleneck'; nevertheless, it outperforms the other two types of models. This suggests that the way a sequence of input tokens `folds' the task image is more important than its length.   

    Concerning the masking schemes, masking only the query panel (Query) throughout the entire training process turns out to be very ineffective. In contrast, Random masking performs much better. This may seem paradoxical, as making predictions about the query panel is less demanding for the learner: as it is located at the end of the third row and the third column of the RPM grid, predicting its properties requires only \emph{extrapolation} of the properties observed in the other rows and columns. In contrast, masking random panels involves also making predictions about the middle panels (requiring \emph{interpolation}) and about the first panels (requiring extrapolation in the opposite direction). A model trained in Random mode has to master all these skills, yet it proves better when tested only at the query panel. This shows that forcing the transformer to detect and reason about patterns observed across the entire puzzle helps it generalize better. 

    While training in the Random mode outperforms the query masking mode by a large margin, Table \ref{tab:tokenizers-masking} suggests that even better predictive accuracy can be attained when the former is followed by the latter in training (Combined mode). Focusing on the query panel in the later stages of training is thus beneficial. 
    The learning curves presented in   
    \ifincludeappendix
    Sec.~\ref{sec:learning-curves} 
    \fi 
    of the SM align with this conclusion: the metrics tend to saturate towards the end of the Random phase and experience increase once training switches to the Query phase. 

    To corroborate these observations, in the \emph{Classification} part of Table \ref{tab:tokenizers-masking} we report the values of selected metrics \emph{calculated for the context panels}. As these panels are not masked out, the model can directly observe them, and predicting their properties is much easier as they do not need to be inferred from the logical rules that govern the puzzle. The metrics are thus much better, with some models attaining almost perfect values. Tokenizer type has an opposite impact on classification compared to prediction: Panel models perform the best, presumably because separate tokenization reduces the `cross-talk' between panels. %
    The complete set of metrics for classification are given in Table 
    \ifincludeappendix
    \ref{tab:tokenizers-masking-context}.
    \fi 

    In Table \ref{tab:transformer_flops}, we characterize the sizes and computational requirements of particular models.  The Row model that excels at predictive accuracy is also the smallest and cheapest at querying. The slight differences in the number of parameters of transformers result from the number of tokens, which determines the number of entries in the learnable embedding used for positional encoding. Relative differences in the total number of parameters are somewhat larger, and stem from the combined dimensionalities of the chunks of output tokens; a chunk is mapped to a property vector using a dense layer and thus its size impacts the number of parameters. Despite these differences in the number of parameters being moderate, the computational cost of querying the Row model is several times lower than for Panel and Task tokenizers. This is due to the larger number of tokens processed by transformers in those models, which leads to quadratically more query-key interactions.

\begin{table}
\caption{The sizes and querying costs for particular models.}\label{tab:transformer_flops}
\centering
\begin{tabular}{lcccc}
\toprule
Tokenizer  &  \multicolumn{2}{c}{\#parameters [M]} &  \multicolumn{2}{c}{MFLOPS} \\
\cmidrule(lr){2-3} \cmidrule(lr){4-5} 
&  Transformer &  Total &  Transformer  &  Total \\
\midrule
Panel &                    2.65 &             11.45 &               87.77 &       2326.16 \\
Task    &                    2.65 &             11.19 &               52.84 &       2002.38 \\
Row  &                    2.64 &             10.68 &               29.02 &        793.97 \\
\bottomrule
\end{tabular}
\end{table}

    \textbf{Does transformer matter?} 
    One of the research questions of this study concerns the importance of the transformer blueprint, i.e.~whether learning to model direct interactions between tokens representing parts of the input brings any advantage compared to more straightforward approaches. To verify this hypothesis, we consider baseline \emph{denseformers} architectures in which the transformer is replaced with a dense subnetwork: the tokens produced by the tokenizer are concatenated into a vector and fed into a subnetwork comprising 5 dense layers of the same size. The output of the last dense layer is then passed to property predictor and undergoes further processing, as in our model, i.e.~it is sliced into 9 chunks used to predict the properties of individual panels (Section \ref{sec:prop-predictor}). There are no other differences between denseformers and our models. 
    
    Given that the Row tokenizer proved most capable (Table \ref{tab:tokenizers-masking}), we design two comparable denseformer variants, with dense layer size 336 and 512, so that the total number of parameters and cost of querying are similar (Tables \ref{tab:densformer_flops} and \ref{tab:transformer_flops}). 
    Each variant is trained in Random masking mode, once with and once without regularization consisting of layer normalization \cite{Ba2016} and dropout.     

\begin{table}
\caption{The sizes and querying costs for dense models.}\label{tab:densformer_flops}
\centering
\begin{tabular}{lcccc}
\toprule
Dense   &  \multicolumn{2}{c}{\#parameters [M]} &  \multicolumn{2}{c}{MFLOPS} \\
\cmidrule(lr){2-3} \cmidrule(lr){4-5} 
layer size &  Transformer &  Total &  Transformer  &  Total \\
\midrule
336        &   2.67 &             10.70 &                5.34 &        770.23 \\
512        &   4.33 &             12.37 &                8.67 &        773.57 \\
\bottomrule
\end{tabular}
\end{table}

     Table \ref{tab:densformer} summarizes the performance of denseformers in terms of metrics from Table \ref{tab:tokenizers-masking}. The densformers are clearly inferior to the transformers: except for the AvgH metric, none of them attains even the worst value of the corresponding metric for the transformers. While layer normalization \cite{Ba2016}  has a positive impact on predictive accuracy, increasing the layer size from 336 to 512 improves the accuracy only slightly, which suggests that boosting it further, beyond 512 units, is unlikely to lead to significant improvements. We thus conclude that the `cross-talk' between tokens representing parts of the RPM task, facilitated by the transformer architecture, brings significant added value, and perhaps is even essential for this kind of tasks. 
 
\begin{table}
\centering
\caption{Comparison of densformer models.}\label{tab:densformer}\begin{tabular}{llrrrrHH}
\toprule
Dense& Layer &  CorrRate &  PropRate &  AvgProp &  AvgH &  Normalized  Hamming &  Hamming \\
size & normal.  \\
\midrule
336 & No &          0.24 &      48.41 &     43.08 &         6.03 &                56.92 &    68.00 \\
    & Yes &          0.45 &      55.19 &     51.24 &         5.43 &                48.76 &    73.08 \\
\midrule
512 & No &          0.28 &      49.76 &     44.76 &         5.88 &                55.24 &    71.59 \\
    & Yes &          0.88 &      55.92 &     52.03 &         5.36 &                47.97 &    74.80 \\
\bottomrule
\end{tabular}
\end{table}

    \textbf{The structure of errors}. 
    We encode the appearance properties as unordered categorical variables, but in fact they are ordinal. In Sec.
    \ifincludeappendix
    ~\ref{sec:errors},
    \fi 
    we show that models are much more likely to commit small errors on these properties than large ones, which implies that they correctly discovered the ordinal nature of attributes, even though it was not engraved in their architectures nor conveyed to them explicitly in training. Such insights are not available in approaches that directly learn to choose answer panels. 
    
    \section{Results: choice tasks}\label{sec:exp-choice}

    In this section, we use the models trained for property prediction in Sec.~\ref{sec:exp-prop} for solving RPM tasks. To this aim, we devise the Direct Choice Maker algorithm (DCM) that makes decisions by comparing the \emph{prediction} for the masked query panel (given context panels) with the \emph{classification} of individual answer panels in the same context. Given a trained model $P$ and a task $T$, DCM proceeds as follows:
    \begin{enumerate}
        \item $P$ is queried on $T$ as in property prediction, i.e., on the context panels of $T$, with the query panel masked out. The 9th property vector $p$ produced in response, corresponding to the query panel, is the model's \textbf{prediction} of the answer to the task. 
        \item For each of the 8 answer panels, $i=1,\ldots,8$, $P$ is queried on $T$ with the query panel replaced with the $i$th answer panel. In each of those queries, the 9th property vector is stored as $p_i$. This will be referred to as \textbf{classification} of a panel (in terms of its properties).  
        \item A distance function $d$ is applied to the pairs $(p, p_i)$, and the answer panel with the minimal $d(p, p_i)$ is returned as the solution to $T$. The distance functions (explained in the following) take into account only the relevant properties, where their relevance is determined by $p_i$ (see Sec.  
        \ifincludeappendix
    ~\ref{sec:relevance-sources}).
    \fi  We use $p_i$ as the source of relevance when calculating $d(p,p_i)$, because classification is in general easier than prediction (cf. Table \ref{tab:tokenizers-masking}), so $p_i$s are less likely to make mistakes in determining the relevance of properties.  
    \end{enumerate}
    We devise three performance metrics, each calculating the percentage of tasks for which DCM selects the correct answer panel. The metrics vary in the type of property vectors (categorical or encoded) and in $d$. \textbf{AccUnique} uses DCM with categorical property vectors and the Hamming distance as $d$. A tie ($d(p,p_i)$ being minimized by two or more answer panels) counts as a failure. \textbf{AccTop} operates like AccUnique, except that a tie on the closest matches counts as a success if one of them points to the correct answer. Finally, \textbf{AccProb} applies DCM to the encoded property vectors, i.e. probability distributions produced by the model, and uses binary cross-entropy for vector elements corresponding to binary properties (e.g., object presence) and categorical cross-entropy for multi-valued properties (e.g. object size), summing them in $d$ over all relevant properties. Similarly to loss functions and metrics used in Sec.~\ref{sec:exp-prop}, all metrics are calculated on the test set.

    \textbf{Optimistic bounds}.\ We first estimate the informal optimistic performance bounds, i.e. the test-set metrics that DCM would attain \emph{if the true property vectors (classifications) were known for answer panels}. These vectors are provided in the RAVEN database, so we use them as $p_i$s in step 2 of DCM (and let them determine the relevance of properties), rather than querying the model. For AccProb, this implies comparing the continuously-valued probabilities produced by the model with one-hot vectors representing the categorical values of true properties.  
    Table \ref{tab:DCM-target-prop} presents the resulting estimates. As expected, AccUnique is the most demanding metric, as it requires the predicted property vector to be strictly closest to the property vector for exactly one of the answer panels. In contrast, AccTop treats ties as successes and thus reports significantly better scores. However, this metric does not reflect the model's capability of pointing to a unique solution among the answer panels. In contrast, AccProb is the most pragmatic metric, due to the low likelihood of ties between answer panels and sensitivity to nuanced, continuously-valued responses of the model, so we focus on this metric in the following.  

\begin{table}
\centering
\caption{Optimistic bounds, with DCM relying on \emph{target} property vectors.}
\label{tab:DCM-target-prop}
\begin{tabular}{llrrr}
\toprule
Tokenizer & Masking &  AccProb &  AccTop &  AccUnique \\
\midrule
Panel & Query &          19.00 &        39.48 &        6.16 \\
         & Random &          55.66 &        70.87 &       37.47 \\
         & Combined &          57.57 &        72.34 &       39.27 \\ \midrule
Task & Query &          65.09 &        74.18 &       38.18 \\
         & Random &          94.84 &        96.89 &       90.70 \\
         & Combined &          95.39 &        97.23 &       92.69 \\ \midrule
Row & Query &          66.45 &        72.04 &       36.10 \\
         & Random &          95.49 &        96.99 &       92.62 \\
         & Combined &          95.90 &        97.48 &       94.04 \\
\bottomrule
\end{tabular}
\end{table}

    The relations between the models in Table \ref{tab:DCM-target-prop} correlate with the quality of property prediction (Table \ref{tab:tokenizers-masking}), with the Row tokenizer being on average better than Task and Panel, the Combined masking mode slightly outperforming the Random mode, and the latter one in turn being much better than the Query-only mode. Expectedly, high accuracy of property prediction implies better choice making. 
    
    \textbf{Accuracy}.\ Table \ref{tab:DCM-pred-prop} presents the actual metrics summarizing models' capability of solving RPM choice tasks, i.e. with $p_i$s resulting from the classification of answer panels.  %
    Bar two models, AccProb is noticeably worse than in Table \ref{tab:DCM-target-prop}, which was expected because the classifications $p_i$ of answer panels can now diverge from the true vectors. Indeed, we calculated also the Correct metric (used in Sec.~\ref{sec:exp-prop} for assessing the accuracy of property prediction) on classification alone in this setting, and it amounted to 75.95\% and 58.77\% for respectively Task and Row. This difference is likely the main factor that makes the former model fare much better in Table \ref{tab:DCM-pred-prop}. We hypothesize that the root cause for this difference is that querying the Row models in classification means replacing an entire input channel of the input image (corresponding to the query panel) with an answer panel, while in training that panel was continuously providing the `neutral' values from a learnable mask. For the Task tokenizer, this affects only $\sfrac{1}{9}$ of the input raster of the entire task (recall that tokens' receptive fields capture the entire input image in all tokenizers).

\begin{table*}
\centering
\caption{Accuracy on choice tasks, based on \emph{predicted} property vectors.}
\label{tab:DCM-pred-prop}
\begin{tabular}{llrrrr}
\toprule
Tokenizer & Masking &  AccProb & AccProb$_{(n=3)}$ & AccTop &  AccUnique \\
\midrule
Panel & Query &    17.79 & --- & 39.13 &       6.33 \\
         & Random &    41.39 & --- &   59.75 &      25.65 \\
         & Combined &    46.85 & --- &  63.82 &      30.74 \\
         \midrule
Task & Query &     5.44 & --- &  42.72 &       0.96 \\
         & Random &    96.33 & 95.81$\pm$1.47 &   96.22 &      83.00 \\
         & Combined &    96.97 & 96.45$\pm$1.13 &   96.57 &      86.30 \\
         \midrule
Row & Query &    30.97 & --- &  63.27 &       5.32 \\
         & Random &    79.23 & 80.58$\pm$5.67 &   94.68 &      25.47 \\
         & Combined &    82.84 & 84.66$\pm$6.28 &   95.43 &      33.53 \\
\bottomrule
\end{tabular}
\end{table*}

    The models trained in the Query masking mode fare the worst again; clearly, the low capacity of predicting properties (Table \ref{tab:tokenizers-masking}) prevents them from choosing the right panels with DCM. For the Panel and Task tokenizer, the pattern is consistent with previous tables: the Combined mode performs better than Random. 
        
    Due to the high computational cost of training, a single model was trained per configuration. To establish statistical significance, we conducted additional runs for the best-performing configurations and report the resulting averages and .95-confidence intervals for sample size 3 obtained in this way in the AccProb$_{(n=3)}$ column of Table \ref{tab:DCM-pred-prop}. The figures largely confirm our earlier observations.

\begin{table}
\centering
\caption{Accuracy on the test set of I-RAVEN benchmark.}
\label{tab:DCM-pred-prop-iraven}
\begin{tabular}{llrrr}

\toprule
Tokenizer & Masking &  AccProb &  AccTop &  AccUnique \\

\midrule
Panel & Query &    45.44 &   64.89 &      27.81 \\
         & Random &    53.59 &   70.22 &      34.61 \\
         & Combined &    60.16 &   74.08 &      41.24 \\
         \midrule
Task & Query &    12.34 &   51.68 &       2.50 \\
         & Random &    94.90 &   95.42 &      86.43 \\
         & Combined &    95.39 &   95.61 &      88.95 \\
         \midrule
Row & Query &    51.35 &   76.74 &      14.30 \\
         & Random &    86.35 &   95.10 &      42.41 \\
         & Combined &    88.52 &   95.55 &      52.09 \\
\bottomrule
\end{tabular}

\end{table}

    In testing, the models are queried on `completed' tasks, with all nine panels present and no panel masked out. In training, they perform classification for the eight unmasked panels and prediction for the single masked panel, but they are never asked to perform classification for all panels. 
    This may be particularly relevant for the Task tokenizer, which in training observes a single panel being masked \emph{in each invocation} (in contrast to Panel and Row); in particular, in the Query-only mode, it is always the lower-right panel. This may explain the particularly bad performance of that model, with the AccProb below the 12.5\% achievable with choosing answer panels at random\footnote{Notice, however, that this explanation ignores the cross-talk between tokens that takes place in further processing by the transformer.}. 

    Table \ref{tab:DCM-pred-prop-iraven} summarizes the performance of the same models on the testing part of the I-RAVEN benchmark \cite{Hu2021}\footnote{Earlier published under the name Balanced-RAVEN \cite{Hu2020}.}, which features the same tasks as RAVEN, however with answer panels generated in an unbiased way. Apart from Task+Query and Task+Combined models that fared best on RAVEN and observe slight deterioration, all remaining models perform better on I-RAVEN. Because the context panels and the \emph{correct} answer panels are identical in both benchmarks, so must be the predictions of properties made by the models for them. Therefore, the differences between Tables \ref{tab:DCM-pred-prop} and \ref{tab:DCM-pred-prop-iraven} can be only due to the classifications of the \emph{incorrect} answer panels. Apparently, the unbiased answer panels from I-RAVEN are less likely to result in property vectors that distort the assessment of relative similarities of the answer panels to the predicted answer. 
     
    \textbf{Comparison with state-of-the-art}. 
    Following a recent survey \cite{Malkinski_Mandziuk_2022}, in Table \ref{tab:sota-raven} we reproduce the test-set accuracy of five RPM solvers reported in past literature on the topic, which attain the best performance on the test part of the RAVEN collection.  
    Table \ref{tab:sota-iraven} presents analogous top results for the I-RAVEN benchmark (see the survey for the performance of other, less capable methods). The reported figures should be juxtaposed with the AccProb metric from previous tables. For reference, we quote also the estimated accuracy of the human performance. 
     
    Compared to these approaches, the performance of several variants of our models is very good, with two of them equipped with the Task tokenizer outperforming not only the reported human accuracy on RAVEN \cite{zhang2019raven}, but also \emph{all previously reported methods on this benchmark}, the best of which attained 94.1\% (column AccProb in Table \ref{tab:DCM-pred-prop} vs. Table \ref{tab:sota-raven}). For I-RAVEN, our method beats all-but-one of the SotA methods (Table \ref{tab:DCM-pred-prop-iraven} vs. Table \ref{tab:sota-iraven}).

\begin{table}
\centering
\caption{State-of-the-art results on the RAVEN test set (source: \protect\cite{Malkinski_Mandziuk_2022}).}
\label{tab:sota-raven}
\begin{tabular}{lc}
\toprule
Method & Accuracy \\
\midrule
Rel-Base \cite{Spratley2020} & 91.7 \\
CoPINet + AL \cite{Kim2020} & 93.5 \\
DCNet \cite{Zhuo2021} & 93.6 \\
CoPINet + ACL \cite{Kim2020} & 93.7 \\
Rel-AIR \cite{Spratley2020}  & 94.1 \\
Ours & 97.0 \\
\midrule
Human \cite{zhang2019raven} & 84.4 \\
\bottomrule
\end{tabular}    
\end{table}

\begin{table}
\centering
\caption{State-of-the-art results on the I-RAVEN test set (source: \protect\cite{Malkinski_Mandziuk_2022}).}\label{tab:sota-iraven}
\begin{tabular}{p{6cm}c}
\toprule
Method & Accuracy \\
\midrule
SRAN \cite{Hu2021} & 60.8 \\
SRAN MLCL+DA\\ \hfill\cite{Malkinski2022} & 73.3 \\
MRNet \cite{Benny2021} & 86.8 \\
SCL \cite{Wu2020} & 95.0 \\
SCL MLCL+DA\\ \hfill\cite{Malkinski2022} & 96.8 \\
Ours & 95.4 \\
\bottomrule
\end{tabular}    
\end{table}

\begin{figure*}
        \begin{center}

            \ifamcs
\else
            \includegraphics[width=0.7\textwidth]{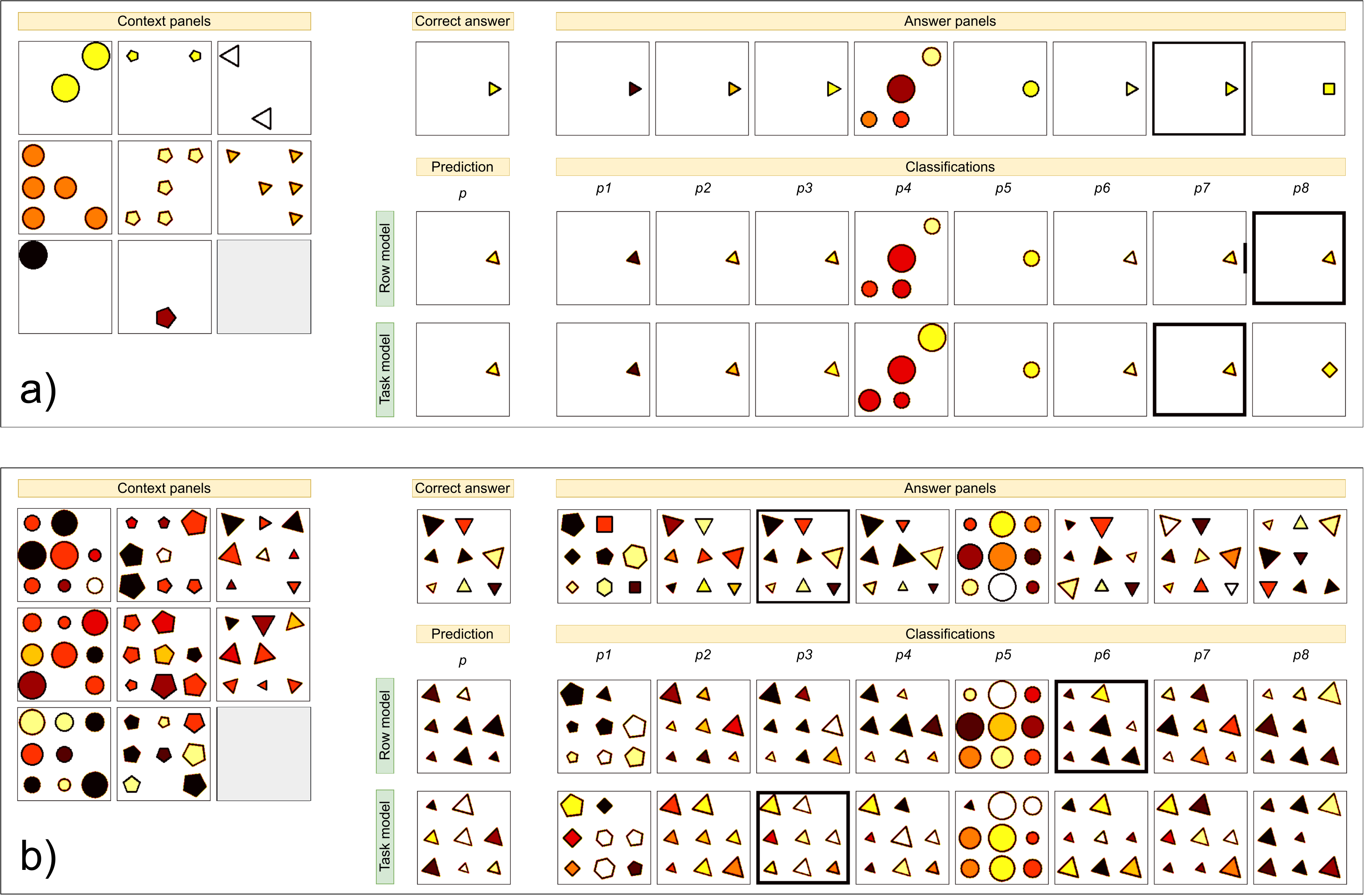}
            \fi

        \end{center}
        \vspace{-3mm}\caption{Solving two RPM tasks (a and b) with Task and Row models (both trained in Combined masking mode). \textbf{Left:} the task. \textbf{Middle:} the correct answer and rendering of models' predictions $p$ (Step 1 of DCM) for the Row and Task model. \textbf{Right:} answer panels and the renderings of classifications $p_i$ generated by Row and Task model (Step 2 of the DCM). The panels corresponding to the most similar property vectors marked with thicker borders. Predictions and classifications are rendered from property vectors produced by the model while fixing rotation angles, as the angle was irrelevant in these tasks. To facilitate analysis, we render the \emph{color} property using pseudocoloring (it is conventionally rendered in grayscale). See Figs.
         \ifincludeappendix
        \ref{fig:additional-vis-1}--\ref{fig:additional-vis-3} for more examples. 
        \fi}
        \label{fig:vis-DCM} 
    \end{figure*}

    \textbf{Examples}. Figure \ref{fig:vis-DCM} compares visually the behavior of Task and Row model for two tasks from the RAVEN test set (rotation angle is fixed when rendering models' predictions and classifications). In the first example (Fig. \ref{fig:vis-DCM}a), both models produce perfect predictions and similar, though imperfect, classifications of answer panels. However, the Row model fails to choose the correct answer, as it classifies the square in the answer panel as a triangle. As a consequence, $p_8$ is more similar to prediction $p$ than $p_7$ (although $p_7$ and $p_8$ look identical when rendered as images, the raw outputs of models varied when assessed with cross-entropy that DCM uses as the distance function $d$). The Task model produces a more faithful classification $p_8$ of the last answer panel and thus correctly points to $p_7$. 
    
    The second task (Fig. \ref{fig:vis-DCM}b) is harder by involving more objects and more complex rules. As a result, not only the classifications, but also the predictions are far from perfect, with imprecise predictions for sizes, colors, and occasionally even shapes of objects (object presence is always correctly predicted and reproduced). Nevertheless, the Task model is more consistent when predicting and classifying and thus chooses the correct answer.  

    More examples are provided in 
    \ifincludeappendix
    Sec.~\ref{sec:additional-vis}
    \fi 
    in the Supplementary Material.

    \textbf{Discussion}.     
    The ultimate superiority of the Task models (Tables \ref{tab:DCM-pred-prop} and \ref{tab:DCM-pred-prop-iraven}) suggests that it is desirable to tokenize tasks as they appear so that the transformer can detect and learn the RPM patterns on its own. Manual engineering of representation, attempted here by the Row tokenizer that directly juxtaposes panels as image channels, does not necessarily help -- even though Row models ranked top at predicting properties of masked panels, they underperformed at classifying answer panels.  

    This study demonstrated that rephrasing a learning task in a multi-dimensional, multi-label fashion can be beneficial for generalization. RPM tasks are in a sense `closed', as the space of responses expected from the model is narrowed down to 8 provided answer panels. Compared to that, learning to classify the properties of visible panels and to predict the properties of masked-out panels is more open-ended, and in a sense generative. It forces the model to derive more detailed patterns from the data and, consequently, leads to better generalization.  
    Moreover, by predicting properties, one may mitigate the biases inadvertently introduced in benchmarks (Table \ref{tab:DCM-pred-prop-iraven}). Last but not least, tracing the process of classifying and predicting properties provides interesting insights (Fig. \ref{fig:vis-DCM}).   

    In mapping an image to a sequence of tokens, the models considered here form an interesting middle ground between purely symbolic approaches and conventional deep learning, in a spirit similar to the Vision Transformer architecture proposed in \cite{Dosovitskiy2020} and neuro-symbolic systems. It is interesting to see that the transformer blueprint is helpful also when approaching a problem that is more abstract than conventional image classification. As evidenced in the presented results (in particular by the failure of the denseformer models; Sec.~\ref{sec:exp-prop}), explicit `perceptual chunking' of representation provided by tokenization and the subsequent contextual reasoning realized with query-key interactions in the transformer allow learning the abstract patterns necessary to predict the missing panel and determine the right answer panel.

    \section{Conclusion and future work}\label{sec:conclusion}

    We have shown that the proposed approach of solving RPM tasks by learning to predict the properties of panels outperforms state-of-the-art models trained to choose answer panels and avoids the biases present in training data. The models fare well despite flattening the 2D structure of the puzzle and can be inspected to a greater extent than end-to-end neural models. In future research, we will consider making the choice makers trainable alongside the model, to allow them to adapt to the deficiencies of classification (identified in Sec.~\ref{sec:exp-choice} and exemplified in Fig. \ref{fig:vis-DCM}) and so enable further improvements. 

    The explicit partitioning of the inference process into property prediction and choice of an answer panel with DCM can be seen as a special case of \emph{task decomposition}, with the properties predicted and classified in the first stage acting as subgoals. In this study, we exploited the subgoals available in the RAVEN benchmark. Prospectively, it would be interesting to synthesize subgoals automatically.  

    \bibliography{main}

\ifincludeappendix
\input{appendix}

\fi

\end{document}

%% file: appendix.tex
    \clearpage

    \setcounter{section}{0}
    \renewcommand{\thesection}{SM\arabic{section}}
    
    \setcounter{table}{0}
    \renewcommand{\thetable}{SM\arabic{table}}

    \setcounter{figure}{0}
    \renewcommand{\thefigure}{SM\arabic{figure}}

    \section*{Supplementary Material}\label{appendix}
    \ifamcs
    \fi

    \section{Technical parameters}\label{sec:technical-params}

    Our model comprises an image tokenizer, transformer, and property predictor (Fig. \ref{fig:method} and Sec. \ref{sec:prop-pred}). 

    \textbf{Tokenizer}. The tokenizer converts panels into a sequence of tokens using a convolutional neural network (CNN), which, depending on the type of tokenizer (described in Sec. \ref{sec:tokenizer}) is applied to the images of individual RPM panels (Panel tokenizer), the images of an entire RPM task (Task tokenizer) or images of rows of RPM task stacked channel-wise (Row tokenizer). We use EfficientNetV2B0 \cite{tan2021} pre-trained on the ImageNet database \cite{deng2009imagenet} as the CNN, but without the final stack of fully connected layers, i.e., only the convolutional stack is used. The superpixels produced by the last layer have 1280 channels. We project them to 128 dimensions using an additional linear layer (a $1\times 1$ convolution) so that 128 is the dimensionality of tokens processed by the transformer.

    \textbf{Transformer.} We use the transformer \cite{Vaswani2017} block architecture from the BERT language model \cite{Devlin2018} (i.e. comprising encoders only), and stack four of such blocks on top of each other. Each block comprises 8 independent transformer heads working in parallel. In each head, the input dimensionality of the transformer is 128, i.e. the queries, keys, and values in the self-attention part of the head are 128-dimensional vectors. The raw tokens produced by the head are processed by a small dense feed-forwarded network with 128 inputs, the first layer of 512 units equipped with the GELU activation function (the \emph{inner size} parameter), and the second linear layer comprising 128 units. The model learns positional embeddings which are added to encoded tokens. The hyperparameters of the transformer are listed in Table \ref{tab:hyperparams}. The dimensionality of the output tokens is 128. 

    \textbf{Predictor.} The predictor maps the chunks of output tokens produced by the tokenizer to a vector of concatenated probability distributions that predict the individual properties of panels. In each invocation, the predictor is applied to a single chunk of concatenated tokens and predicts the properties of a single panel. The number of tokens in a chunk varies by the type of tokenizer; therefore, the input dimensionality $k$ of the predictor is a multiple of the dimensionality of the transformer's output tokens (128) and amounts to:
    \begin{itemize}
        \item 1152 for the Panel tokenizer (9 tokens per chunk $\times$ 128), 
        \item 897 for the Task tokenizer (7 tokens per chunk $\times$ 128), 
        \item 384 for the Row tokenizer (3 tokens per chunk $\times$ 128).    
    \end{itemize}
    The predictor consists of two fully connected layers, each comprising 1000 units (the first of them having the input dimensionality $k$) and using the GELU non-linearity \cite{Hendrycks2016}, along with an output layer consisting of 557 neurons (which is the combined dimensionality of the above-mentioned probability distributions. The units corresponding to encodings of individual properties share a softmax activation function (see Sec. \ref{sec:property-vectors} for details), except for present$_i$ properties that encode the presence/absence of objects in slots, and are thus binary -- for those properties, we use the sigmoid activation function. Before each layer, there is layer normalization \cite{Ba2016} with an epsilon value of 0.001. \hfill$\square$
 
    To train the model, we use the Adam optimizer with a learning rate of 0.001 and a batch size of 64.
    
\begin{table}[b]
\centering
\caption{Hyperparameters of the transformer.}\label{tab:hyperparams}
\begin{tabular}{lr}
\toprule
Hyperparameter & Value \\
\midrule
Number of blocks        &       4 \\
Number of heads     &       8 \\
Size (dimensionality of tokens) &     128 \\
Inner size       &    512 \\
Activation       &    GELU \\
Dropout          &    0.1 \\
Layer Normalization &  0.001 \\
\bottomrule
\end{tabular}
\end{table}

    \section{Determining the relevance of properties (sources of relevance)}\label{sec:relevance-sources}
    
    When applying the loss function or a metric to a pair of property vectors $p_1$ and $p_2$, one of them is appointed as the source of relevance, i.e.\ it determines which of the elements of the compared vectors should be taken into account. More specifically, 
    \begin{itemize}
        \item For the loss function, the source of relevance is the target vector, 
        \item When evaluating a model with a metric, the source of relevance is the target vector, %
        \item In DCM, the source of relevance is the vector of classified properties $p_i$ (see the beginning of Sec. \ref{sec:exp-choice}). 
    \end{itemize}
    The relevance of individual elements of the compared property vectors is determined in accordance with the hierarchy of properties presented in Sec. \ref{sec:property-vectors}, in the following order:
    \begin{itemize}
        \item The value of the \emph{arrangement} variable determines the subset $V$ of \emph{present}$_i$ variables that are relevant (out of the total of 25 slots), 
        \item The \emph{present}$_i$ binary variables determine the relevance of individual object slots in $V$, narrowing so $V$ down to $V' \subseteq V$, 
        \item $V'$ determines the relevance of corresponding appearance properties (of the total of 75) describing the objects in each slot.
    \end{itemize}

    \section{Loss function}\label{sec:loss-function}
    
    When calculating the loss function applied to models in training and the $d()$ metric used in the DCM (Sec. \ref{sec:exp-choice}; except for the case when $d()$ is the Hamming distance), contributions of individual properties are multiplied by weights determined empirically at the beginning of this project. These are 1.0 for \emph{arrangement}, 2.83426987 for \emph{present}$_i$, and 0.85212836, 1.096005, and 1.21943385 for respectively color, size, and shape. The weights were determined so that the average contributions of particular types of properties (arrangement, presence, color, shape, size) to the distance function were equal. This was done for one of the first models obtained, and the weights remained fixed in all experiments.

    \section{Variance and stability of results}

    The relatively high computational costs of training and limited access to computational resources did not allow performing multiple runs for all configurations considered in the paper -- except for the top performing ones, for which we report the statistics over multiple runs in the AccProb$_{(n=3)}$ column of Table \ref{tab:DCM-pred-prop}. 
    Nevertheless, our models are largely insensitive to random initialization. For instance, in preliminary experiments, the standard deviation of AccProb varied from $0.23$ to $0.26$ percent point for a $6$-run trial, which translates respectively into $.95$-confidence intervals of $\pm 0.21$ and $\pm 0.19$ percent points. Given the relatively large differences between the compared models, this level of randomness has no significant impact on the conclusions drawn in the paper.

    \section{Technical implementation}

    The software framework for conducting experiments has been written in Python ver.\ 3.10, with the deep learning models implemented in the TensorFlow library ver.\ 2.11. The source code and documentation of the framework is available at \url{https://github.com/jakubkwiatkowski/abstract_compositional_transformer}. 

    Models have been trained on Tesla V100 GPU card and servers equipped with Xeon Gold 5115 processor and 16GB of RAM. A typical training cycle of a single model lasted 15 hours. Querying a model to solve a single task on this hardware takes $\sim$2 %
    seconds.

   \begin{figure*}
        \centering
        \includegraphics[width=0.3\textwidth]{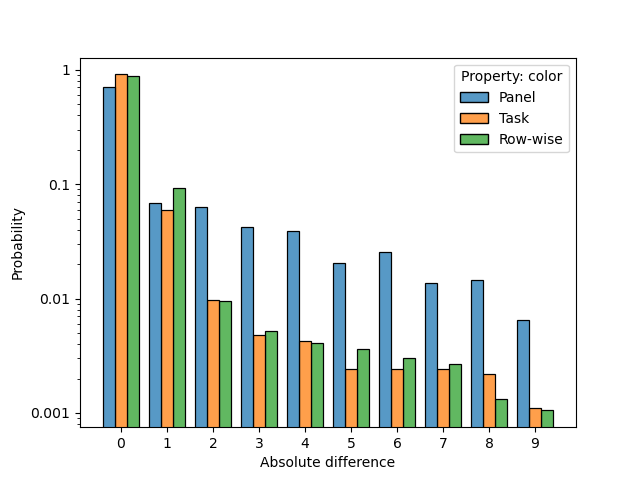}
        \includegraphics[width=0.3\textwidth]{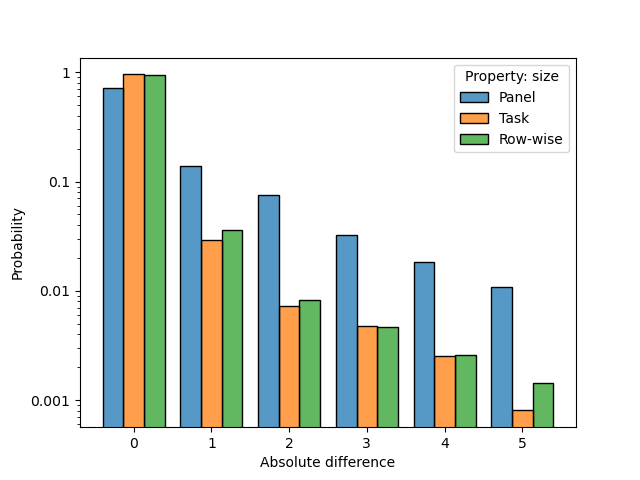}
        \includegraphics[width=0.3\textwidth]{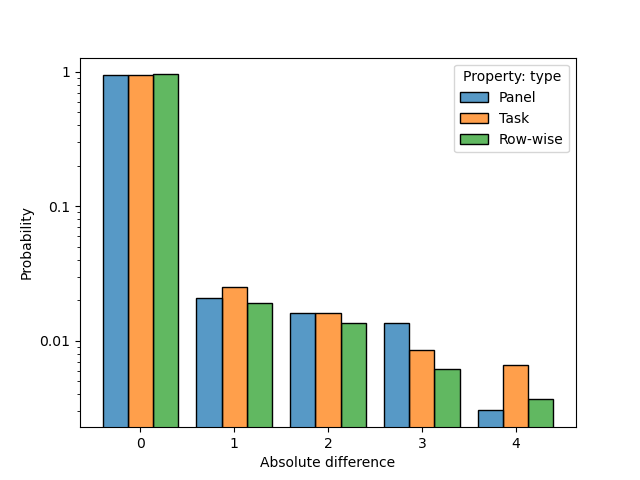}
        \caption{Histograms of absolute difference errors committed by the models on object properties. Analogous distributions for untrained (random) models (not shown here for clarity) are very close to uniform. }\label{fig:abs-diff-hist}
    \end{figure*}

    \section{Accuracy of classification for the visible panels}

    Table \ref{tab:tokenizers-masking-context} presents the quality of predictions provided by particular models when evaluated on the context panels, i.e. classifying the properties of visible panels. As in all other tables presented in this study, these values have been calculated from the test set. 
    
\begin{table}
\centering
\caption{Comparison of configurations \emph{on the context panels}.}
\label{tab:tokenizers-masking-context}
\begin{tabular}{llrrrrHH} %
\toprule
Tokenizer & Masking &   CorrRate &  PropRate &  AvgProp &  AvgH &  Normalized  Hamming &  Hamming \\
\midrule
Panel & Query &         97.54 &      99.53 &     99.50 &         0.05 &                 0.50 &    65.02 \\
         & Random &         98.38 &      99.49 &     99.44 &         0.05 &                 0.56 &    64.65 \\
         & Combined &         98.28 &      99.53 &     99.49 &         0.05 &                 0.51 &    64.84 \\ \midrule
Task & Query &         89.69 &      98.28 &     98.35 &         0.31 &                 1.65 &    67.26 \\
         & Random &         88.44 &      97.90 &     98.00 &         0.37 &                 2.00 &    67.46 \\
         & Combined &         88.33 &      97.87 &     97.98 &         0.38 &                 2.02 &    67.61 \\ \midrule
Row & Query &         84.03 &      97.70 &     97.68 &         0.37 &                 2.32 &    64.50 \\
         & Random &         87.64 &      98.17 &     98.22 &         0.31 &                 1.78 &    67.65 \\
         & Combined &         87.85 &      98.20 &     98.25 &         0.31 &                 1.75 &    67.25 \\
\bottomrule
\end{tabular}
\end{table}

    \section{Analysis of the structure of errors}\label{sec:errors}

    While RPM is formally an \emph{abstract} reasoning task, its representation is visual. It becomes thus interesting to investigate more closely how the models interpreted the visual features of objects present in the panels. One insight that can be elicited here concerns the structure of errors committed on predicting the properties of objects, i.e. color, size, and type. The former two have domains that are naturally ordered;  the values of the type property can be ordered by the number of edges of the shape: triangle, square, pentagon, hexagon and circle. This allows us to define the metric of absolute difference between the index of the predicted position in the property's domain, and the index of the actual position. 
    
    In Fig. \ref{fig:abs-diff-hist}, we plot the histograms of absolute difference error (MAE on the above-mentioned indices), calculated on the test set, for the models trained in the Combined mode. These models are quite good at predicting individual properties (see PropRate in Table \ref{tab:tokenizers-masking}), hence they are very likely to achieve the absolute difference of 0 (notice the logarithmic scale of the probability axis)\footnote{PropRate in Table \ref{tab:tokenizers-masking} captures all properties, including arrangement and object presence, while Fig. \ref{fig:abs-diff-hist} summarizes the \emph{object} properties only.}.     
    More interestingly, the probability of committing error decreases monotonically almost everywhere with increasing values of absolute difference, %
    despite the models being trained to predict these properties as categorical probability distributions (one-hot encoding). In other words, the ordinal nature of these properties was not revealed to the models in training, nor explicitly engraved in their architectures. Yet, the errors clearly relate to the visual properties of objects. For instance, if the actual object to be predicted is a square, a model will be on average more likely to predict a triangle or a pentagon, than a hexagon or a circle. 
    
    This clearly proves that the models managed to meaningfully relate the low-level visual features extracted by image tokenizers to abstract high-level properties of objects -- even though this capability is only a part of the overall abstract reasoning task. We hypothesize thus that the models are likely to generalize well beyond the considered domain, like recognizing an octagon as a new shape category that is an 'interpolation' between hexagon and circle, or extrapolating size by being able to properly operate when objects are smaller than 0.4 or greater than 0.9 on the scale of the size property (these numbers being the actual range of object size in the RAVEN family of benchmarks). These characteristics hold for all three properties and all models illustrated here; however, the distributions for the non-zero errors are most uniform for the type property, which suggests that discovering and capturing the ordinal nature of this attribute is the most challenging.

    \section{Learning curves}\label{sec:learning-curves}

    \begin{figure}
        \begin{center}
        \includegraphics[width=\columnwidth]{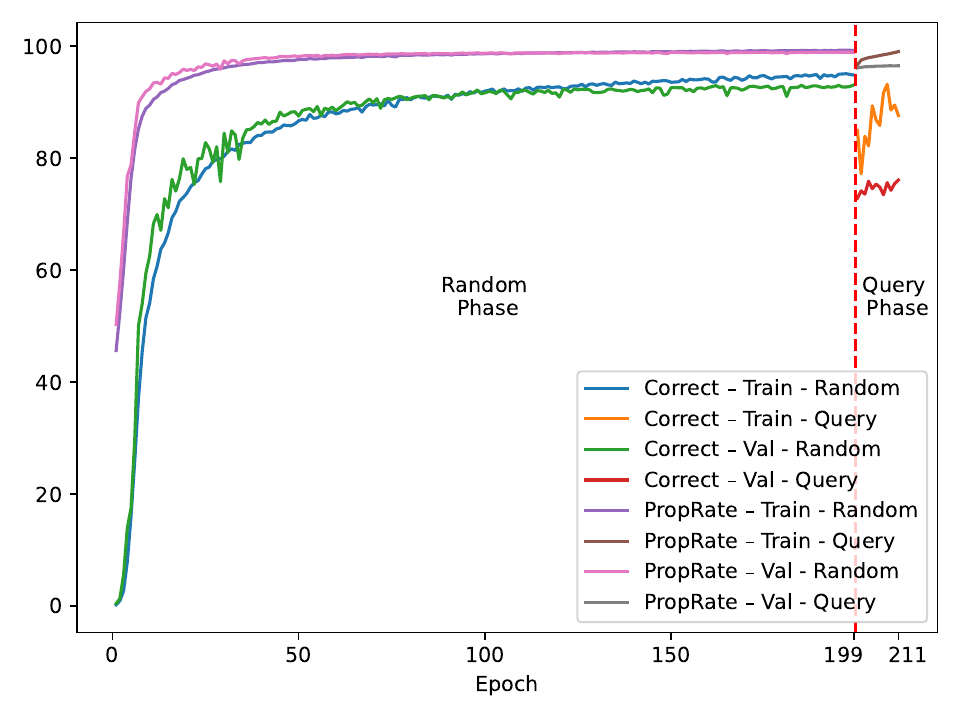}
        \end{center}
        \caption{The learning curve in terms of Correct and AvgProp metrics for the Task tokenizer model trained in Combined masking mode, with the Random masking phase terminated by the early stopping condition in 199th epoch and the Query masking phase terminated after 12 epochs.}
        \label{fig:learning-curve} 
    \end{figure}
    \begin{figure}
        \begin{center}
        \includegraphics[width=\columnwidth]{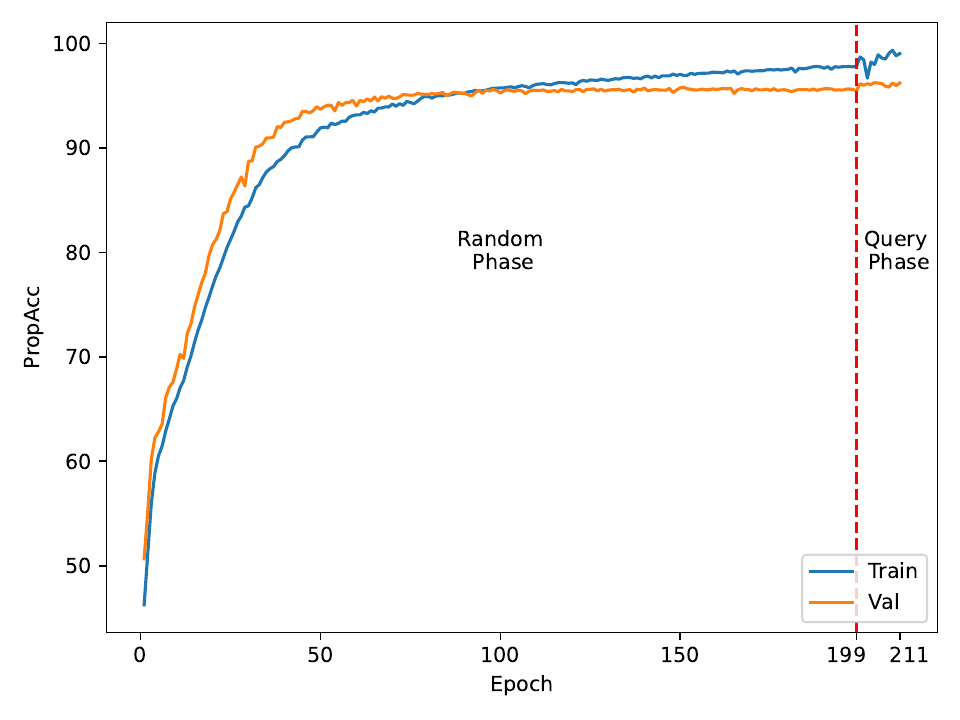}
        \end{center}
        \caption{The learning curve for the same run as in Fig.~\ref{fig:learning-curve}, in terms of the PropAcc metric.}
        \label{fig:learning-curve-propAcc} 
    \end{figure}

    Figure \ref{fig:learning-curve} presents how the Correct and AvgProp metrics change in training on the training and validation set, for the model equipped with Task tokenizer trained in the Combined masking mode (cf. Sec.~\ref{sec:exp-prop}. In this particular run, learning in the Random phase has been terminated by the validation set-based stopping condition in the 199th epoch (out of 200 provided as the maximum budget). For those first 199 epochs, we plot the combined metrics on classification and prediction (i.e. for both unmasked and masked panels). The final part of the plot (the Query phase) captures the metrics for prediction only, hence the sudden drop is only apparent because prediction is harder than classification. That part corroborates the usefulness of the Combined mode, where the model elaborated further improvements, e.g. by $\sim$3 percent points on the Correct metric (Table \ref{tab:tokenizers-masking}). In this particular run, the Query masking phase was terminated in the 12th epoch by the validation set-based stopping condition.
    
    Figure \ref{fig:learning-curve-propAcc} presents an analogous plot for the same run, but for \emph{PropAcc}, the metric that was not used in the main body of the paper for brevity. \emph{PropAcc} is the average accuracy on individual properties: arrangement accuracy, position accuracy, type accuracy, size accuracy, and color accuracy. This metric resembles AvgProp; however, differences arise due to relevance: panels vary in the number of relevant objects and properties, which causes these metrics not being equivalent. For this plot, \emph{PropAcc} has been calculated only for the currently masked panel. Also for this metric, the transition from the Random mode to Query mode is marked with observable improvement. Interestingly, these curves are also less steep than those in Fig.~\ref{fig:learning-curve} in the initial part of the run: by being calculated only for the masked panels, it reflects only the prediction capability, which is harder than the classification.

\section{Examples of biased answer sets in RAVEN}\label{sec:biased-answer-set}

\begin{figure}
    \begin{center}
    \includegraphics[width=\columnwidth]{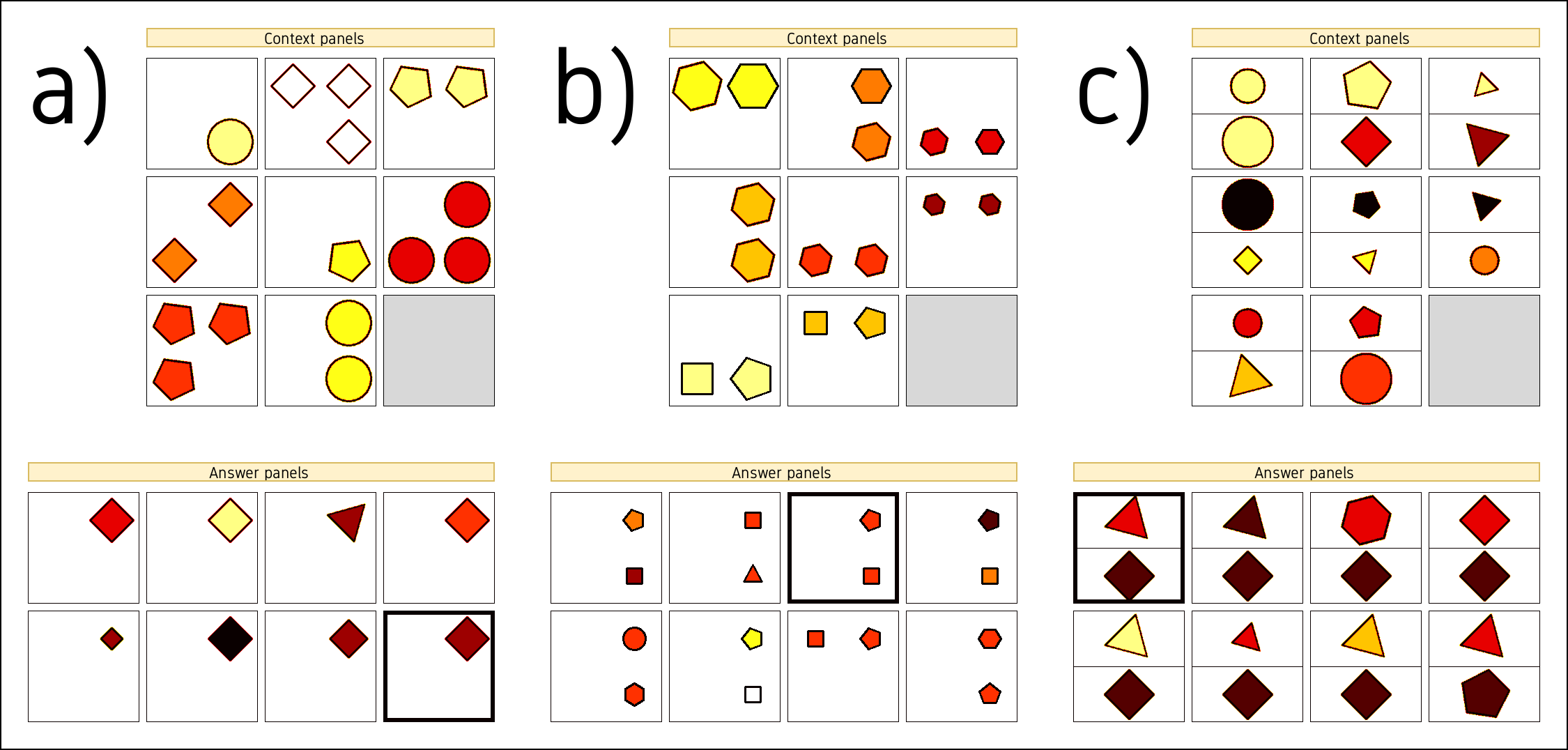}
    \end{center}
    \caption{Examples of tasks with biased answer sets from the RAVEN collection: the correct answer can be determined by selecting the panel featuring the most common attributes across all answer panels (see the text for details).}
    \label{fig:biased-answer-sets} 
\end{figure}

In the Introduction, we pointed to the flaws in the design of some of the tasks in the RAVEN collection. In this section, we illustrate this problem with three tasks shown in Fig.\ \ref{fig:biased-answer-sets}. 

In the example a), the statistics of the occurrence of properties in the answer panels are as follows: 
\begin{itemize}
    \item shape: 7 squares and 1 triangle, 
    \item color: 4 crimson figures, 1 yellow, 1 black, 1 red, and 1 coral, 
    \item size: 7 big figures and 1 small.
\end{itemize}
By selecting the most frequently occurring properties, we obtain a query panel with a big crimson square in the upper-right corner, which is the correct answer in this puzzle. 

Similar reasoning can be applied to more complex tasks with more figures, as demonstrated in example b). We observe that red is the most frequently used color in the answer set, the most common pair of figures is a pentagon and a square, and 7 out of 8 times the location of figures is the same (at the right edge of the panel). This again leads to easily picking the correct answer, i.e.\ red pentagon and a square at the right-hand edge of the puzzle. 

Analogous reasoning reveals the flaws in the answer set in example c), allowing one to easily pick the correct answer without even looking at the context panels. In other words, looking for modal values in the marginal distributions of object properties of answer panels greatly facilitates (if not makes trivial) solving quite many tasks from the RAVEN benchmark.  

\section{Additional visualizations}\label{sec:additional-vis}

\begin{figure*}
        \centering
        \includegraphics[width=0.8\textwidth]{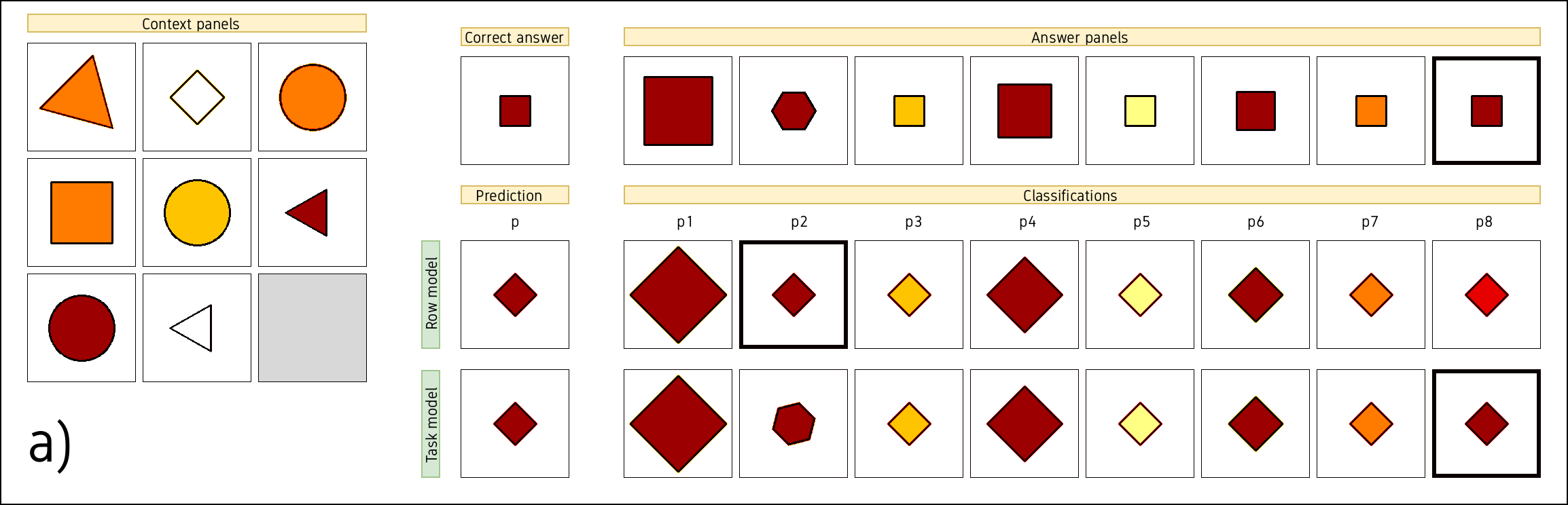}
        \includegraphics[width=0.8\textwidth]{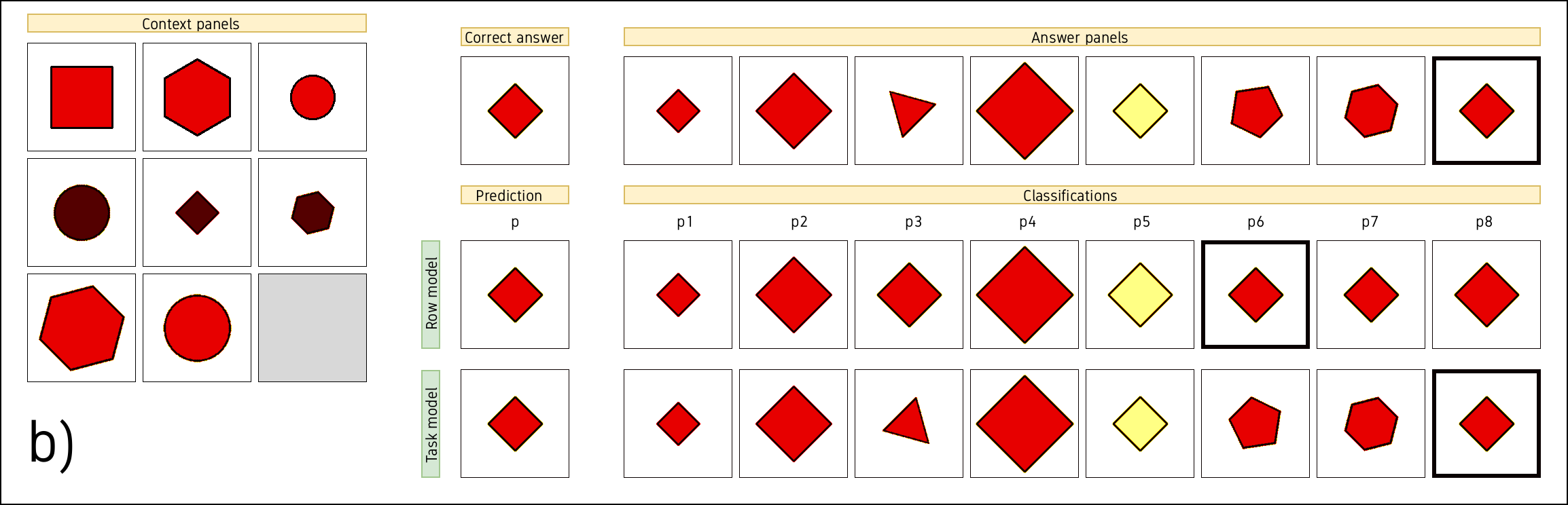}
        \includegraphics[width=0.8\textwidth]{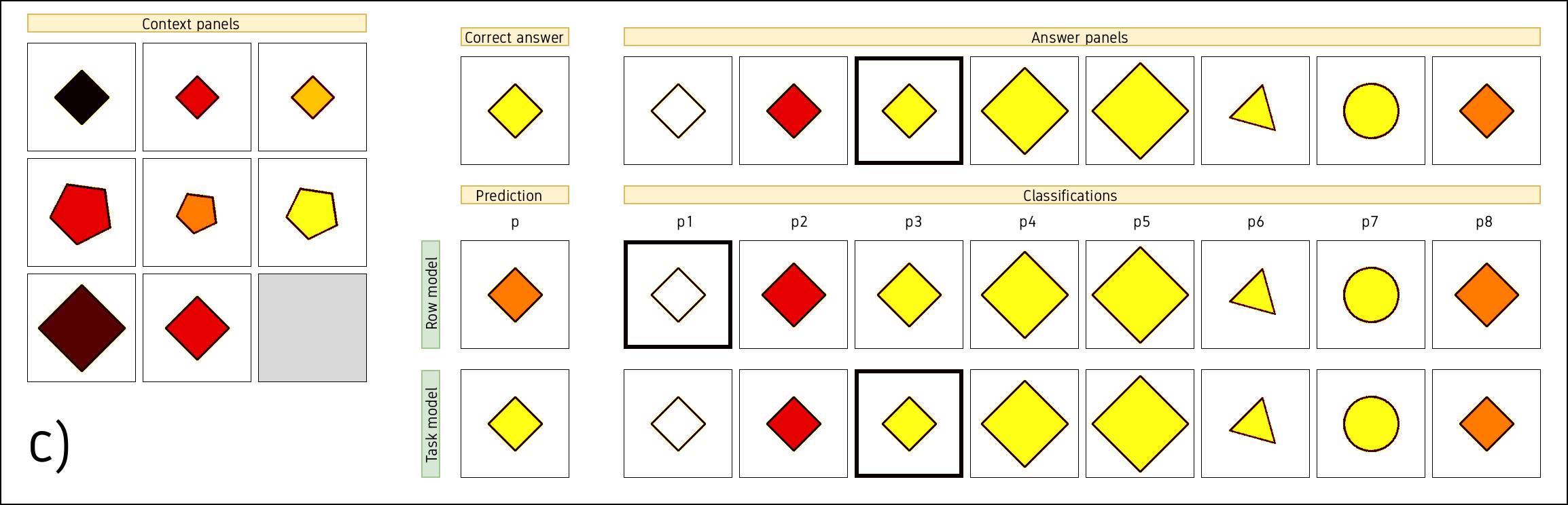}
        \includegraphics[width=0.8\textwidth]{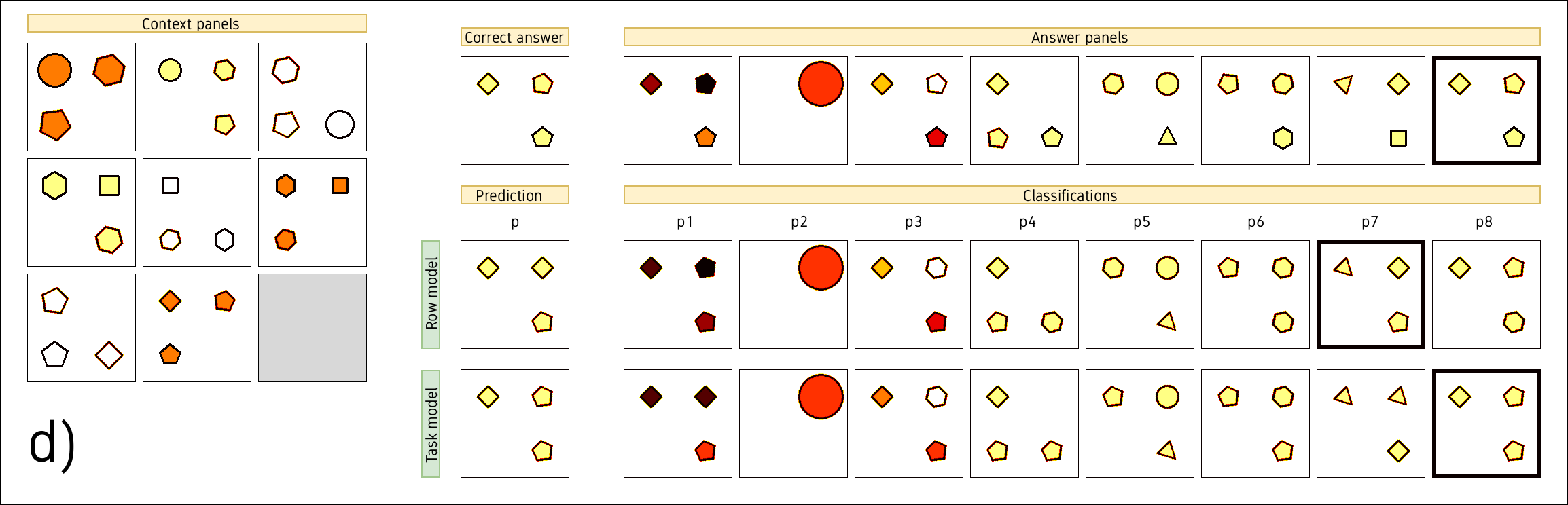}
        \includegraphics[width=0.8\textwidth]{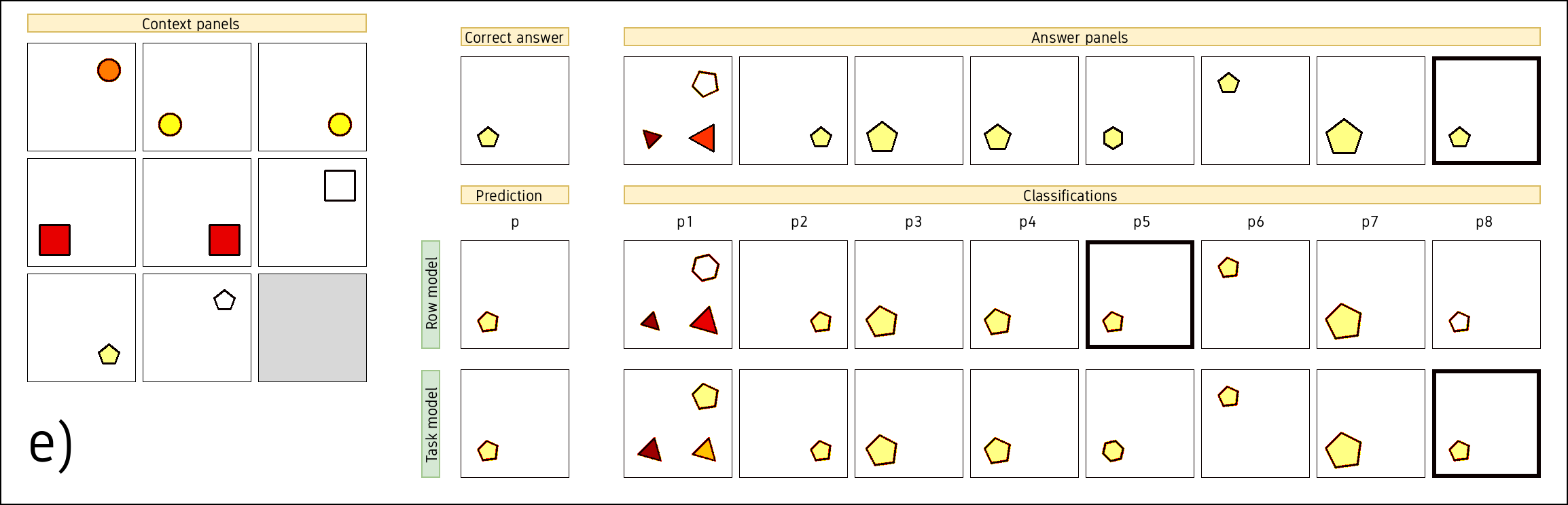}
        \caption{Visual comparison of the behaviors of the Row and Task models (trained in the Combined mode) on additional tasks from the RAVEN benchmark. The interpretation of each inset is identical to that used in Fig. \ref{fig:vis-DCM} and explained there. }\label{fig:additional-vis-1}
    \end{figure*}

\begin{figure*}
        \centering
        \includegraphics[width=0.8\textwidth]{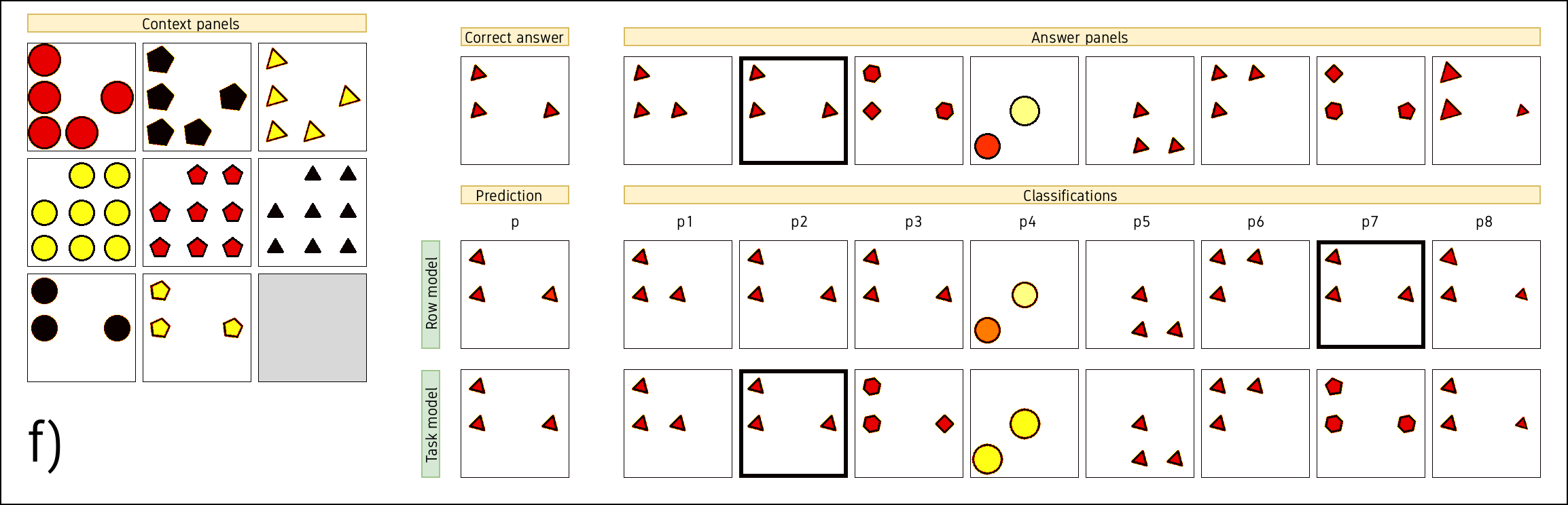}
        \includegraphics[width=0.8\textwidth]{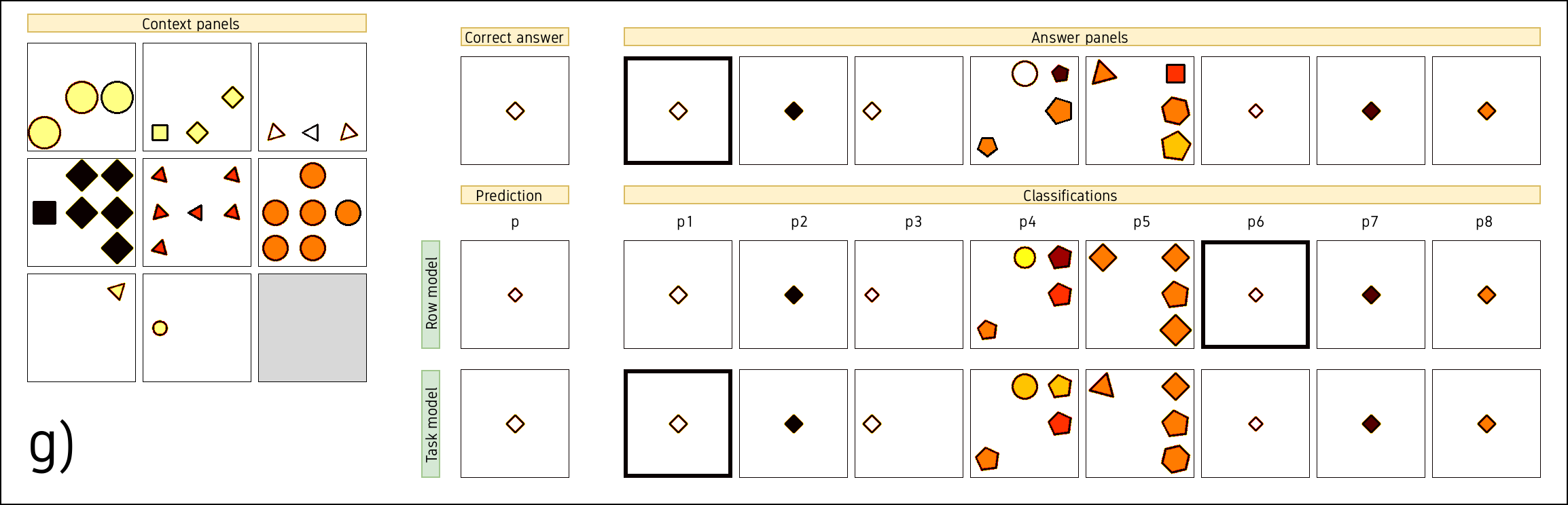}
        \includegraphics[width=0.8\textwidth]{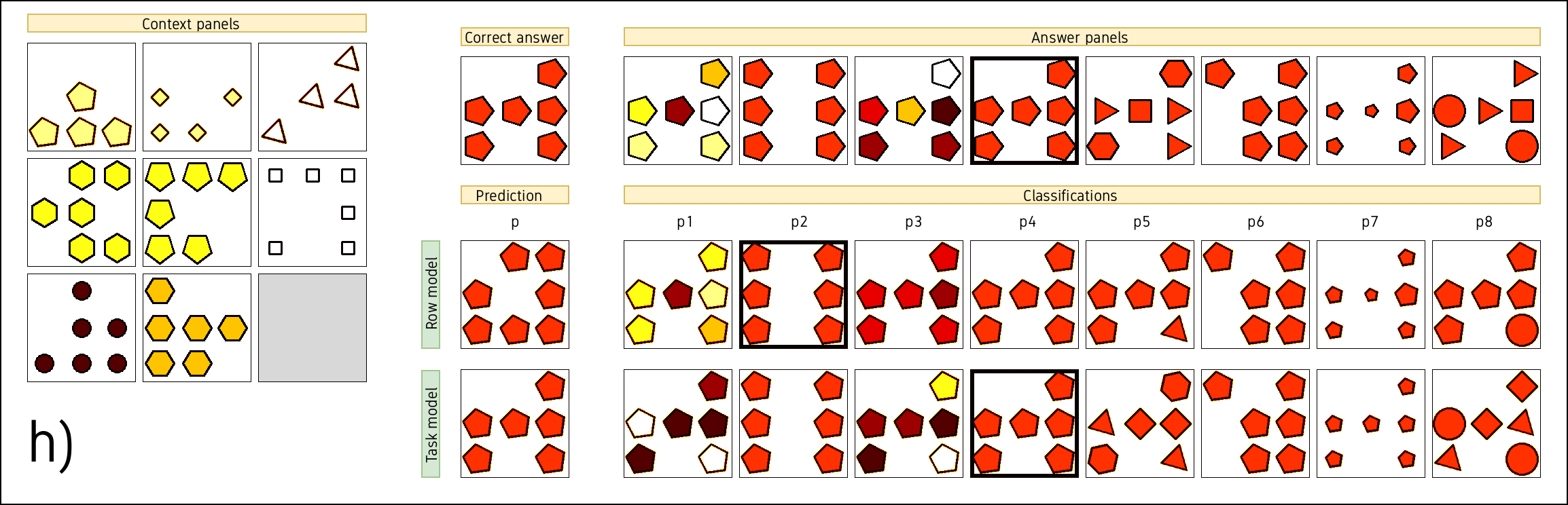}
        \includegraphics[width=0.8\textwidth]{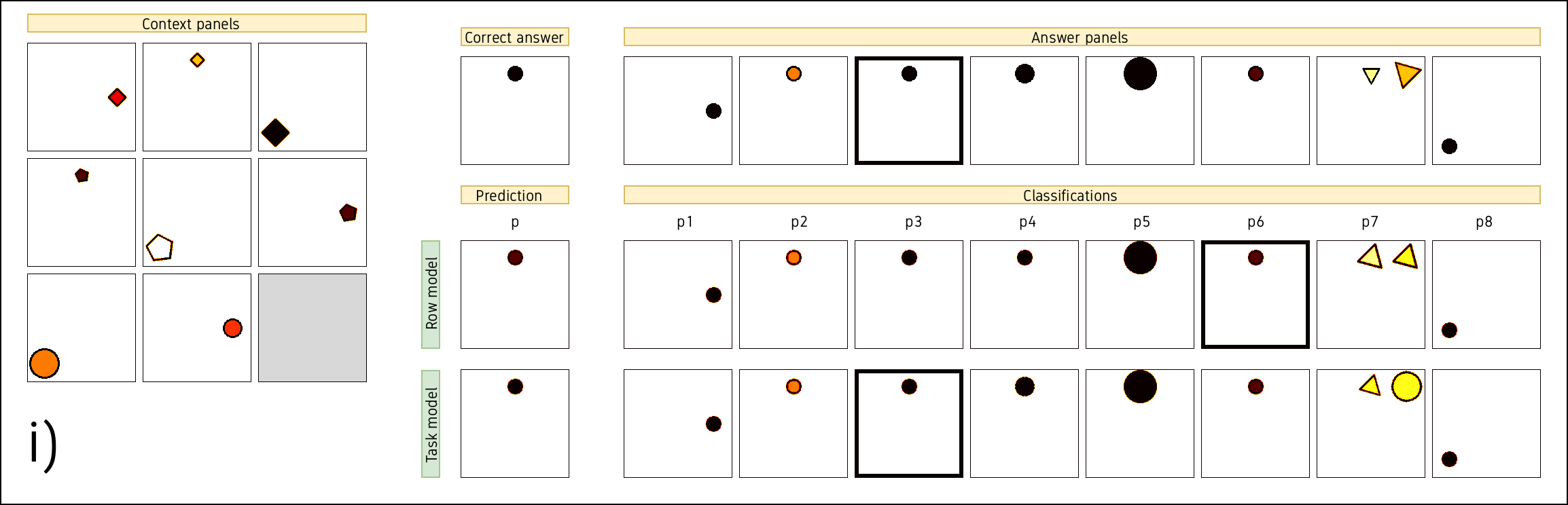}
        \includegraphics[width=0.8\textwidth]{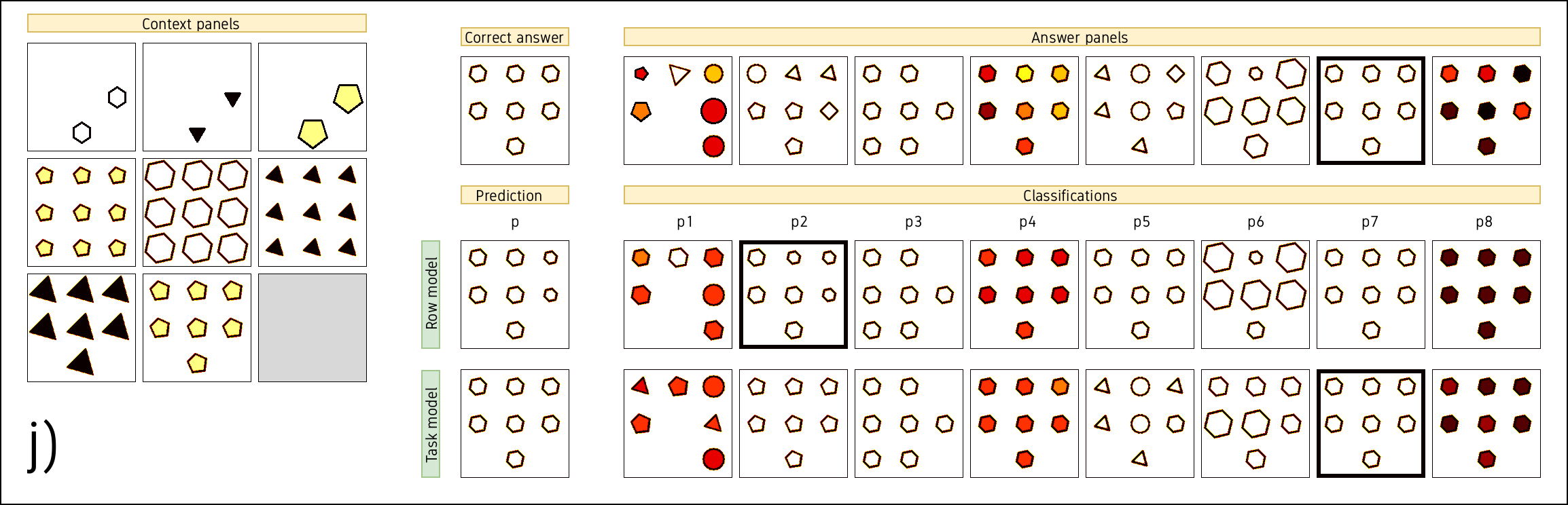}
        \caption{Visual comparison of the behaviors of the Row and Task models (trained in the Combined mode) on additional tasks from the RAVEN benchmark. The interpretation of each inset is identical to that used in Fig. \ref{fig:vis-DCM} and explained there. }\label{fig:additional-vis-2}
    \end{figure*}

\begin{figure*}
        \centering
        \includegraphics[width=0.8\textwidth]{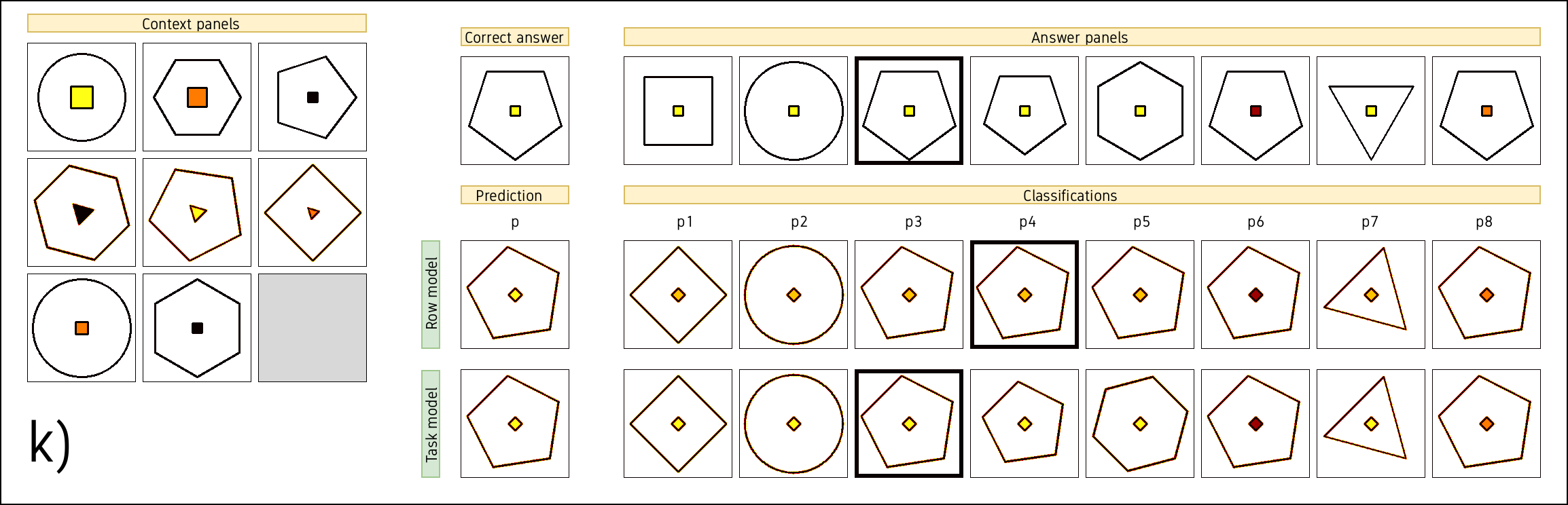}
        \includegraphics[width=0.8\textwidth]{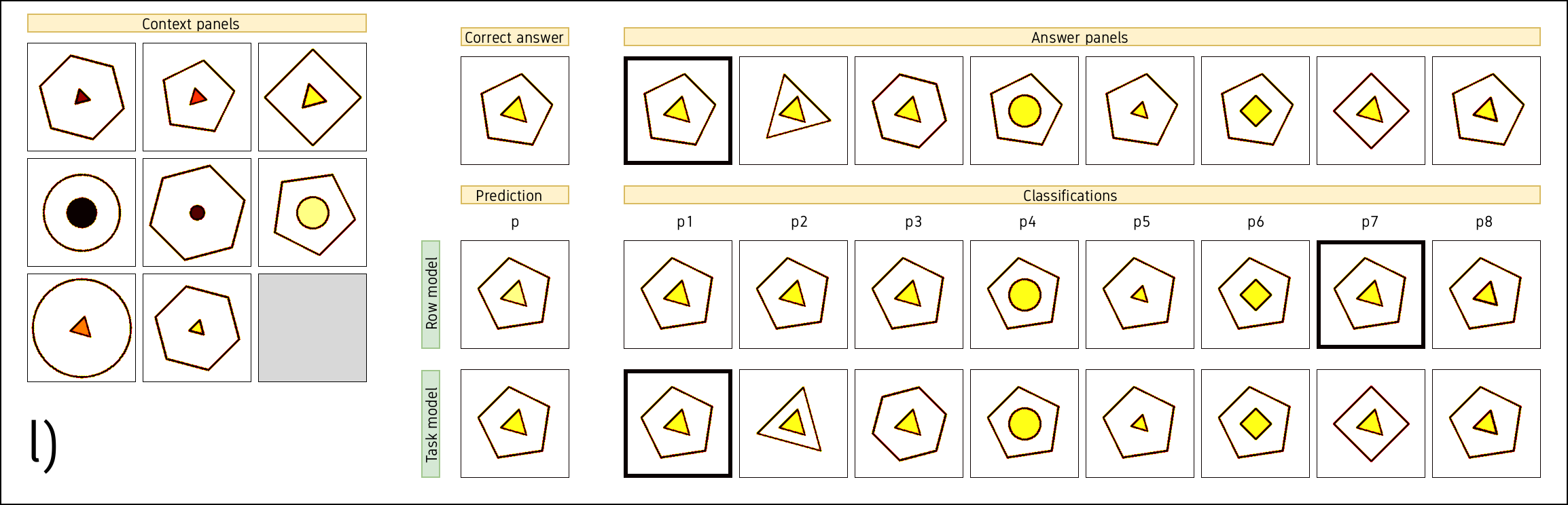}
        \includegraphics[width=0.8\textwidth]{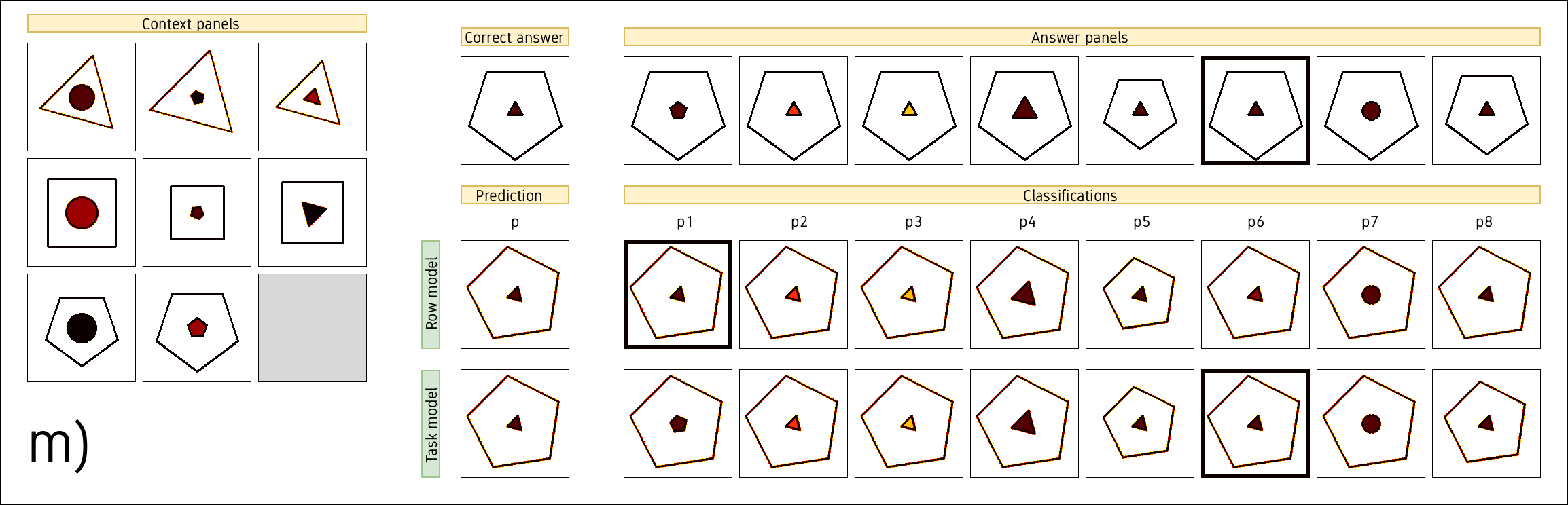}
        \includegraphics[width=0.8\textwidth]{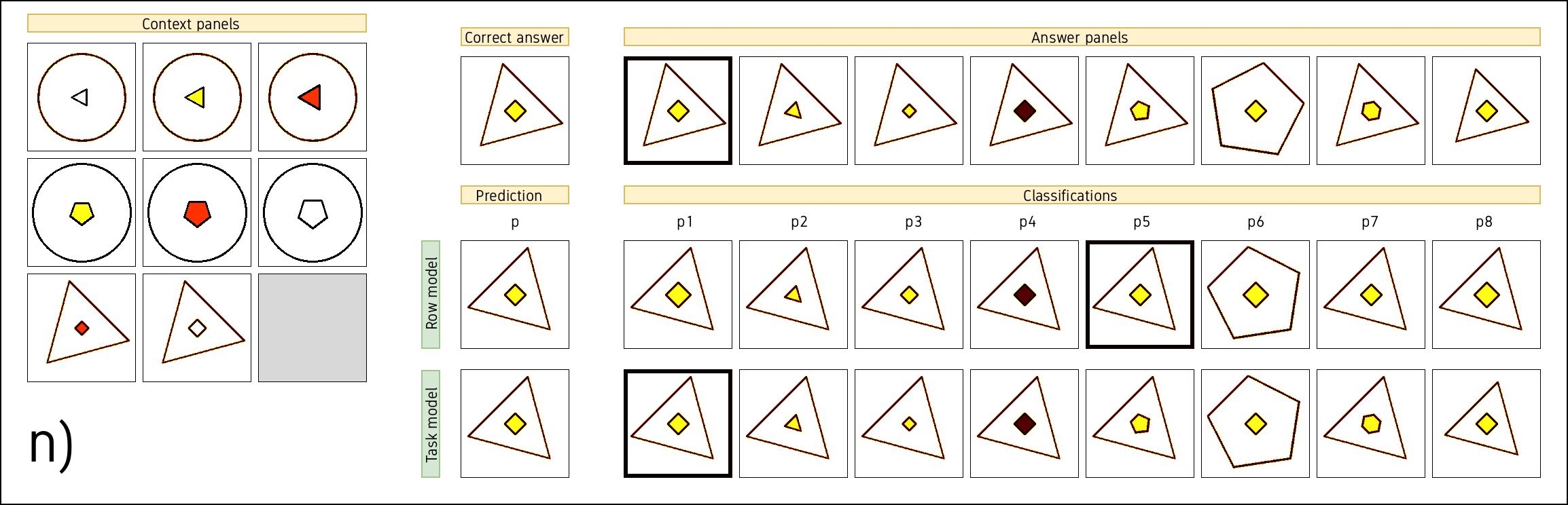}
        \includegraphics[width=0.8\textwidth]{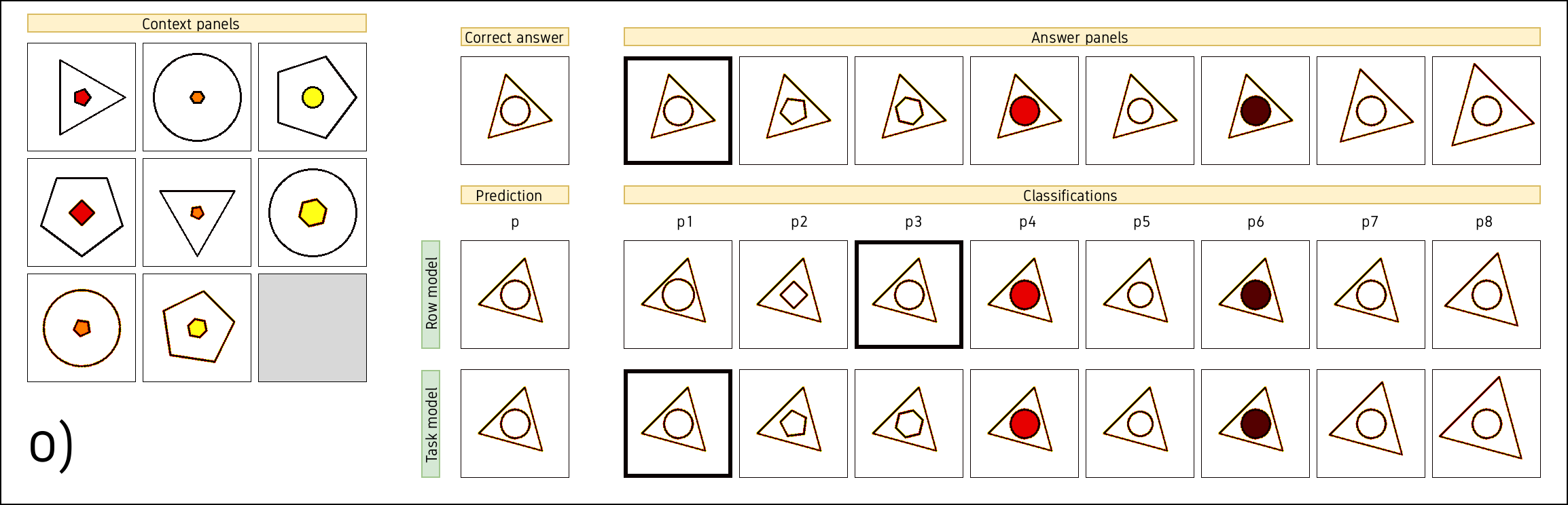}
        \caption{Visual comparison of the behaviors of the Row and Task models (trained in the Combined mode) on additional tasks from the RAVEN benchmark. The interpretation of each inset is identical to that used in Fig. \ref{fig:vis-DCM} and explained there. }\label{fig:additional-vis-3}
    \end{figure*}

    Figures \ref{fig:additional-vis-1}--\ref{fig:additional-vis-3} present the visualizations of the behaviors of the Row and Task models on 15 additional tasks, presented in the same way as in Fig. \ref{fig:vis-DCM}. Both models were trained in Combined masking mode. 
    
    The layout of visualization in each inset is as follows. \textbf{Left:} the task. \textbf{Middle:} the correct answer and the rendering of models' predictions $p$ (Step 1 of DCM) for the Row and Task model. \textbf{Right:} the answer panels and the renderings of classifications $p_i$ generated by the Row and Task model (Step 2 of the DCM). The panels corresponding to the most similar property vectors are marked with thicker borders. Predictions and classifications are rendered from property vectors produced by the model while fixing rotation angles, as the angle is irrelevant in RAVEN tasks. To facilitate analysis, we render the \emph{color} property using pseudocoloring (it is conventionally rendered in grayscale). 

    \clearpage
    \clearpage
